\documentclass{article} 
\usepackage{iclr2026_conference,times}

\usepackage{booktabs} 
\usepackage{multirow} 
\usepackage{graphicx} 
\usepackage{amsmath}  
\usepackage[table]{xcolor} 

\usepackage{graphicx}
\usepackage{subcaption}
\usepackage{caption}

\usepackage{amsmath,amsfonts,bm}

\def\eqref#1{equation~\ref{#1}}

\def\1{\bm{1}}

\DeclareMathAlphabet{\mathsfit}{\encodingdefault}{\sfdefault}{m}{sl}
\SetMathAlphabet{\mathsfit}{bold}{\encodingdefault}{\sfdefault}{bx}{n}

\usepackage{hyperref}
\usepackage{url}

\usepackage[capitalize,noabbrev]{cleveref}
\usepackage{adjustbox}
\usepackage{float}

\usepackage{siunitx}

\usepackage{algorithm}        
\usepackage{algpseudocode}

\usepackage{wrapfig}

\title{\lookaheadkv: Fast and Accurate KV Cache Eviction by Glimpsing into the Future without Generation}

\author{
\vspace{0.1em}
Jinwoo Ahn\thanks{\hspace{.06in}Equal contribution $^{\dagger}$ Corresponding authors.}   \hspace{3mm}  Ingyu Seong$^{*}$  \hspace{2mm} Akhil Kedia  \hspace{2mm} Junhan Kim \hspace{2mm} Hyemi Jang \\
\vspace{0.4em}
\textbf{Kangwook Lee$^{\dagger}$ \hspace{2mm} Yongkweon Jeon$^{\dagger}$} \\
\hspace{0.1mm} Samsung Research \\
\texttt{\{jinwoo.ahn, ingyu.seong, kw.brian.lee, dragwon.jeon\}@samsung.com}
}

\newcommand{\lookaheadkv}{\textsc{LookaheadKV}}

\iclrfinalcopy 
\begin{document}

\maketitle

\begin{abstract}
Transformer-based large language models (LLMs) rely on key–value (KV) caching to avoid redundant computation during autoregressive inference.
While this mechanism greatly improves efficiency, the cache size grows linearly with the input sequence length, quickly becoming a bottleneck for long‑context tasks.
Existing solutions mitigate this problem by evicting prompt KV that are deemed unimportant, guided by estimated importance scores.
Notably, a recent line of work proposes to improve eviction quality by “glimpsing into the future”, in which a draft generator produces a surrogate future response approximating the target model's true response, and this surrogate is subsequently used to estimate the importance of cached KV more accurately. However, these approaches rely on computationally expensive draft generation, which introduces substantial prefilling overhead and limits their practicality in real-world deployment.
To address this challenge, we propose \lookaheadkv, a lightweight eviction framework that leverages the strength of surrogate future response without requiring explicit draft generation.
\lookaheadkv \ augments transformer layers with parameter‑efficient modules trained to predict true importance scores with high accuracy.
Our design ensures negligible runtime overhead comparable to existing inexpensive heuristics, while achieving accuracy superior to more costly approximation methods.
Extensive experiments on long-context understanding benchmarks, across a wide range of models, demonstrate that our method not only outperforms recent competitive baselines in various long-context understanding tasks, but also reduces the eviction cost by up to $14.5$×, leading to significantly faster time-to-first-token.
Our code is available at \url{https://github.com/SamsungLabs/LookaheadKV}.
\end{abstract}
\section{Introduction}\label{sec:intro}
Extending the context length of Large Language Models (LLMs) is becoming increasingly critical for many emerging applications: processing long documents~\citep{bai-etal-2024-longbench, wang-etal-2024-leave, hsieh2024ruler}, repository-level code understanding and generation~\citep{luo2024repoagent, liu2024repobench, jimenez2024swebench}, in-context learning~\citep{li2024long, manyshot_icl}, etc.
However, a central challenge in enabling these applications is that the key‐value (KV) cache size grows linearly in sequence length, which rapidly becomes a bottleneck for inference, restricting scalable deployment of such applications.
For example, even for moderate-sized models, such as LLaMA3.1–70B~\citep{dubey2024llama} in half-precision, storing a single $128$K-token sequence already takes up $40$GB of memory, while scaling to $1$M tokens requires $320$GB, exceeding the memory capacity of high-end consumer hardware.

A growing line of work addresses this challenge by identifying salient tokens to achieve effective KV cache eviction without loss of performance~\citep{li2024snapkv, cai2024pyramidkv, galim2025speckv, wang2025laq, zhang2023h2o}.
Early methods often rely on simple heuristics, in which token importance is estimated based on the self-attention scores of a subset of the input tokens. SnapKV~\citep{li2024snapkv}, for instance, leverages the attention weights between the suffix of the input and the preceding context to estimate the importance of each prompt token.
More recently, several studies~\citep{galim2025speckv, wang2025laq} reveal that leveraging the model's response, rather than the input prompt, can greatly improve the eviction quality.
Furthermore, they show that a low-cost generated draft response (generated using a smaller draft model~\citep{galim2025speckv}, for instance), which closely approximates the true response, can serve as a powerful proxy for accurately estimating the importance scores.

\begin{figure}
    \centering
    \includegraphics[width=0.95\columnwidth,trim={0cm 1.5cm 0cm 0cm},clip]{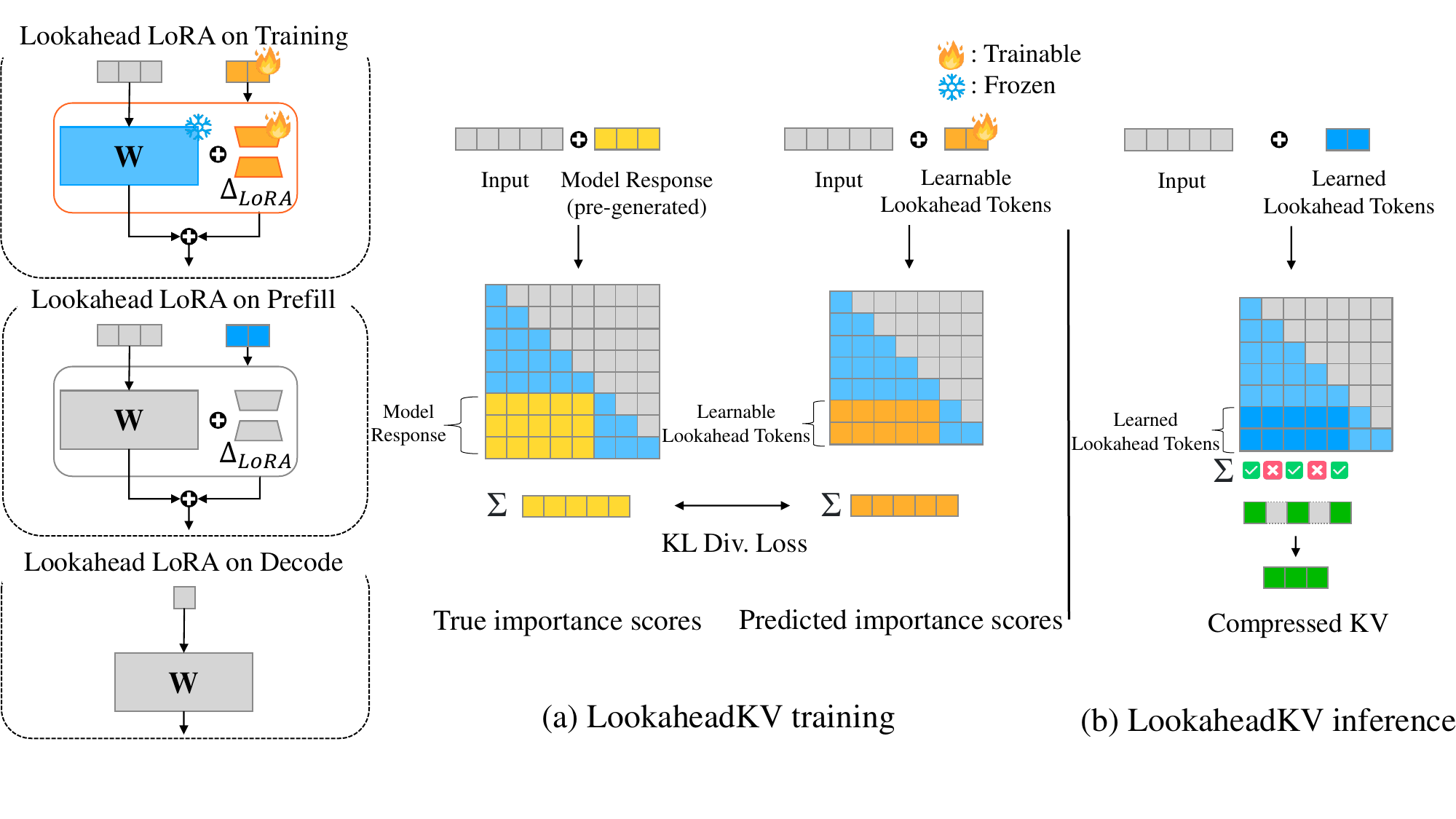}
    \caption{
    \textbf{An overview of \lookaheadkv} \textbf{(a) Training.} During training, lookahead tokens and lookahead LoRA are trained to predict the ground-truth importance scores obtained with pre-generated model response via a KL divergence loss. \textbf{(b) Inference.} During prefill, \lookaheadkv \ utilizes the learned modules to identify essential tokens and compress the KV cache, facilitating memory-efficient decoding.
    }\vspace{-0.3cm}
    \label{fig1:main}
\end{figure}
\begin{wrapfigure}[16]{r}{0.4\textwidth}
\setlength{\intextsep}{0pt}
    \centering
    \includegraphics[width=0.38\textwidth]{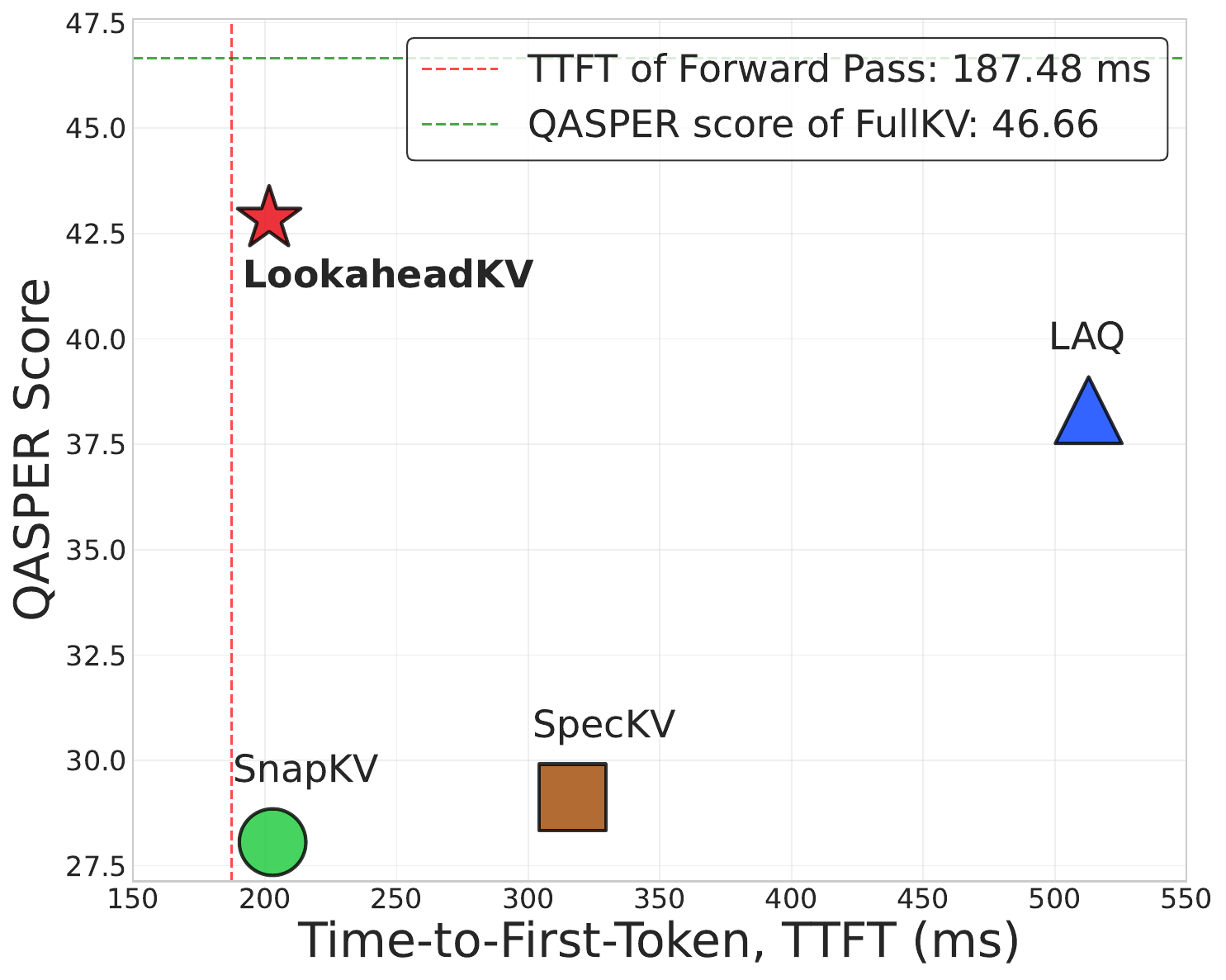}
    \caption{Accuracy-overhead trade-off across KV cache eviction methods.}
    \label{fig2:tradeoff}
\end{wrapfigure} While these draft-based methods substantially improve eviction quality, they still face a trade-off between performance and latency, since their draft token generation step is computationally expensive.
\cref{fig2:tradeoff} presents the trade-off between accuracy and overhead of different approaches using the QASPER benchmark~\citep{dasigi-etal-2021-qasper} and LLaMA3.1-8B-Instruct~\citep{dubey2024llama}.
While simpler approaches like SnapKV induce minimal latency overhead, they suffer severe performance degradation under highly constrained budget settings.
On the other hand, Lookahead Q-Cache (LAQ)~\citep{wang2025laq}, a draft-based approach, shows impressive results even in extremely limited budget settings.
However, this approach incurs prohibitive computational overhead by generating an extra draft response, which limits its practicality in latency-sensitive applications such as mobile devices.

To overcome this limitation, we introduce \lookaheadkv, a novel KV cache eviction method that augments LLMs with parameter-efficient modules, capable of accurately predicting future attention patterns, eliminating the need for costly draft token generation.
As shown in~\cref{fig2:tradeoff}, our method effectively overcomes the accuracy-overhead trade-off, achieving minimal performance loss with negligible overhead.
\lookaheadkv, as depicted in~\cref{fig1:main}, employs a set of learnable special tokens, together with lookahead LoRA modules, novel selectively activated low-rank adapters, to produce queries that can reliably estimate token-importance scores.
By fine-tuning them to predict the true importance scores, \lookaheadkv \ effectively minimizes the quality loss incurred by KV cache eviction with marginal inference overhead.

To rigorously assess the effectiveness of \lookaheadkv, we evaluate it on a diverse set of long‑context benchmarks~\citep{bai-etal-2024-longbench, hsieh2024ruler, ye2025longproc, zheng2023mtbench} across multiple models of varying sizes~\citep{dubey2024llama, yang2025qwen3technicalreport}.
Experimental results consistently demonstrate that \lookaheadkv \ outperforms strong baselines across multiple budgets and context lengths while incurring significantly less eviction latency.

To summarize, our contributions are as follows:
\begin{itemize}
    \item We propose \lookaheadkv, a novel KV cache eviction framework that employs learnable lookahead tokens and special LoRA modules to predict the importance scores from the model's true response without explicitly generating costly approximate response.
    \item Through extensive experiments, we demonstrate that the proposed approach is effective and robust across different models and context lengths. It remains superior in low-budget settings, providing a useful solution in resource-constrained environments.
    \item By conducting a rigorous analysis of eviction latency, both theoretically and empirically, we show that our method incurs negligible eviction overhead of less than $2.16$\% at $32$K context length, which is up to $14.5$$\times$ lower than the overhead incurred by draft-based approaches.
\end{itemize}
\section{Background}\label{sec:preliminary}
The primary objective of the KV cache eviction methods considered in this work, including our proposed approach, is to accurately estimate the importance score of individual key-value pairs of prompt tokens using attention weights, in order to guide the eviction process.
In the following section, we formally define the problem of KV cache eviction and briefly discuss how prior methods have approached it.

\textbf{KV Cache Eviction Using Importance Scores.}
Let \(X = \{x_{1}, ..., x_{n_{\text{in}}}\}\) be an input token sequence (e.g., a user instruction, part of a code snippet, etc.) and \(Y = \{y_{1}, ..., y_{n_{\text{out}}}\}\) the model's generated response to \(X\).
For a given layer and attention head in an LLM, the attention scores of the complete sequence are given by:
\begin{equation}
\mathbf{Q} = \begin{bmatrix} \mathbf{X} \\ \mathbf{Y} \end{bmatrix} \mathbf{W}_{q}
\;\;\;\;\;\;\;\;\;\;\;\;\;\;
\mathbf{K} = \begin{bmatrix} \mathbf{X} \\ \mathbf{Y} \end{bmatrix} \mathbf{W}_{k}
\;\;\;\;\;\;\;\;\;\;\;\;\;\;
\mathbf{A} \;=\;
\operatorname{Softmax}\!\Bigg(
    \frac{\, \mathbf{Q} \, \mathbf{K}^{\top} \,}
         {\sqrt{d}}
\Bigg),\
\end{equation}
where \(\mathbf{X} = [\mathbf{x}_{1}, ..., \mathbf{x}_{n_{\text{in}}}]^{\top} \in \mathbb{R}^{n_{\text{in}} \times d} \) and \(\mathbf{Y} = [\mathbf{y}_{1}, ..., \mathbf{y}_{n_{\text{out}}}]^{\top} \in \mathbb{R}^{n_{\text{out}} \times d} \) are the hidden states of the input prompt and model-generated response, respectively.
For better readability, we omit the layer and head index.
We define the ground-truth importance scores \(\mathbf{s}_{\text{GT}} = [s_1, ..., s_{n_{\text{in}}}]\) of the KV cache as the average cross-attention scores between the queries of \(\mathbf{Y}\) and the keys of \(\mathbf{X}\), i.e., \(s_{j} = \frac{1}{n_{\text{out}}} \sum_{i \,=\, n_{\text{in}} + 1}^{n_{\text{in}} + n_{\text{out}}} \mathbf{A}_{\,i,j}\).
Intuitively, these scores quantify the relative contribution of each prompt token’s key–value pair to the model’s response generation.
Based on these scores, the pruned KV cache can be obtained by retaining a subset of (e.g., TopK) important KV pairs to minimize the attention output perturbation, such that

\(
    \operatorname{Attn}(x, \text{KV}_{\text{orig}}) \approx \operatorname{Attn}(x, \text{KV}_{\text{GT}}),
\)
where \(\text{KV}_{\text{orig}}\) and \(\text{KV}_{\text{GT}}\) are the original and evicted KV cache using the ground-truth importance scores, respectively.

However, since the model's true future response is unknown during the prefill phase, such scores cannot be computed directly.
Consequently, prior methods resorted to constructing a surrogate response sequence \(\tilde{\mathbf{Y}} = [\tilde{y}_{1}, ..., \tilde{y}_{n_{\text{window}}}]^{\top} \in \mathbb{R}^{n_{\text{window}} \times d} \) to approximate the model’s (partial) future response and predict the attention pattern:
\begin{equation}
\tilde{\mathbf{Q}} = \begin{bmatrix} \mathbf{X} \\ \tilde{\mathbf{Y}} \end{bmatrix} \mathbf{W}_{q}
\;\;\;\;\;\;\;\;\;\;\;\;\;\;
\tilde{\mathbf{K}} = \begin{bmatrix} \mathbf{X} \\ \tilde{\mathbf{Y}} \end{bmatrix} \mathbf{W}_{k}
\;\;\;\;\;\;\;\;\;\;\;\;\;\;
\tilde{\mathbf{A}} \;=\;
\operatorname{Softmax}\!\Bigg(
    \frac{\, \tilde{\mathbf{Q}} \, \tilde{\mathbf{K}}^{\top} \,}
         {\sqrt{d}}
\Bigg),\
\end{equation}
resulting in the estimated importance score vector  \(\mathbf{s}_{\text{approx}} = [\tilde{s}_1, ..., \tilde{s}_{n_{\text{in}}}]\) whose entries are computed as \(\tilde{s}_{j} = \frac{1}{n_{\text{window}}} \sum_{i \,=\, n_{\text{in}} + 1}^{n_{\text{in}} + n_{\text{window}}} \tilde{\mathbf{A}}_{\,i,j}\).
In short, these methods aim to obtain the estimated score vector whose ranking is similar to that of the ground-truth, such that the overlap between the retained KV pairs and \(\text{KV}_{\text{GT}}\) is high.
Various approaches have been suggested to approximate the future response for effective KV cache eviction.

\textbf{Prompt-based Approaches.} SnapKV~\citep{li2024snapkv} uses the suffix of input prompt to compute the estimate of the future importance scores.
It has been widely adopted as a simple and effective KV cache eviction method because it can reuse attention weights from the prefill forward pass, requiring only marginal extra computation.

\textbf{Draft-based Approaches.}
Recently, several works have proposed to use a low-cost generator to generate a (partial) approximate response first, and subsequently use it to estimate the future importance scores.
For example, SpecKV~\citep{galim2025speckv} employs a smaller LLM to generate a draft response, while Lookahead Q-Cache (LAQ)~\citep{wang2025laq} first applies SnapKV to the target model to generate a draft response, which is in turn used to approximate the future salience.

\begin{figure}[t]
    \centering 

    \begin{subfigure}[b]{0.48\textwidth}
        \centering
        \includegraphics[width=0.9\textwidth]{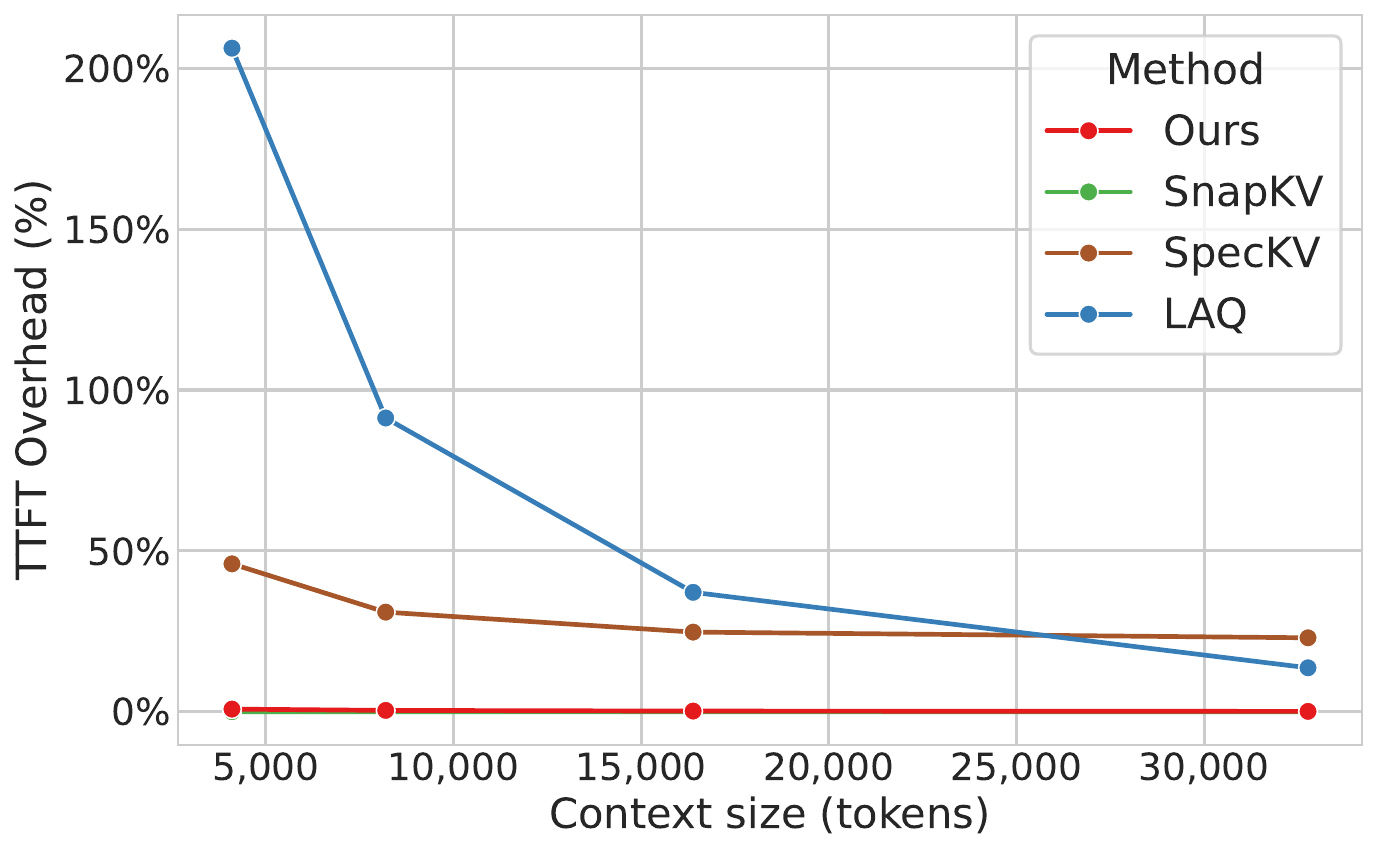}
        \caption{Theoretical latency overhead}
        \label{fig:overhead_pct}
    \end{subfigure}
    \hfill 
    \begin{subfigure}[b]{0.48\textwidth}
        \centering
        \includegraphics[width=0.9\textwidth]{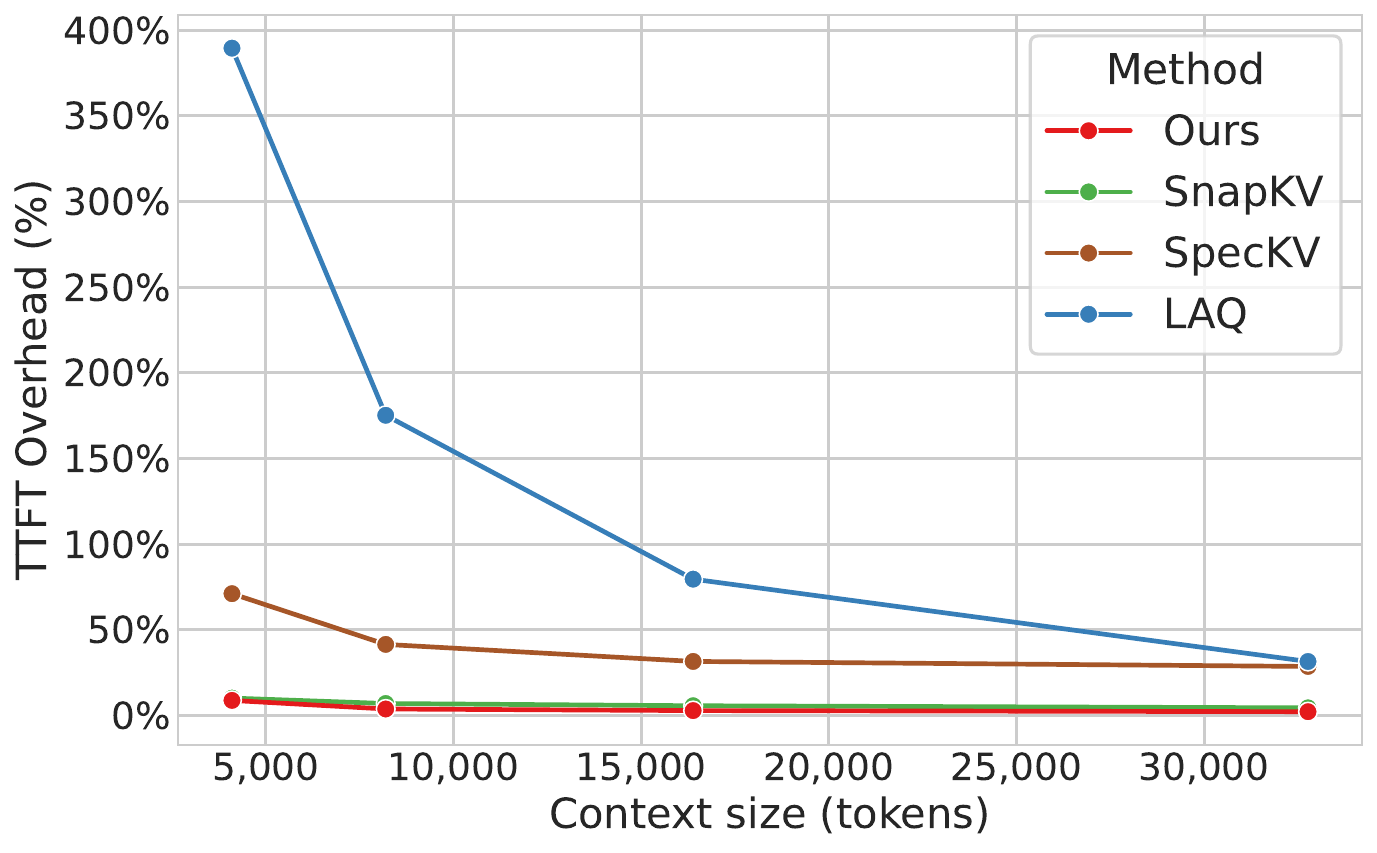}
        \caption{Actual latency overhead}
        \label{fig:practical_overhead_pct}
    \end{subfigure}

    \caption{Time-to-First-Token (TTFT) latency overhead ratio across context lengths.
Similar to SnapKV, \lookaheadkv \ introduces negligible TTFT overhead across all tested context lengths; draft-based methods (LAQ, SpecKV) incur substantial latency, especially for shorter contexts.}
    \label{fig:overhead_comparison_graph}
\end{figure}

These draft-based methods have consistently shown superior performance compared to simple heuristics~\citep{li2024snapkv}, demonstrating the effectiveness of employing surrogate future response.
However, the explicit draft generation step still incurs substantial additional compute, resulting in significant increase in latency, as shown in~\cref{fig:overhead_comparison_graph}.
In summary, existing methods face a clear trade‑off: inexpensive heuristics are fast but less accurate, whereas draft‑based techniques improve performance at the cost of increased inference time.

\section{Proposed Method: \lookaheadkv}
To overcome the challenge of fast and accurate importance prediction, we introduce \lookaheadkv, a framework that augments the LLM with a set of lightweight learnable modules which are optimized to predict ground‑truth importance scores.
\lookaheadkv \ achieves the best of both worlds \textit{by glimpsing into the future without generation:}
\textbf{1)} it eliminates the need for the explicit draft generation step, resulting in significantly faster KV cache eviction.
\textbf{2)} it employs learned special tokens that serve as implicit future response for importance estimation, leveraging the strength of draft-based methods without their computational overhead.

\subsection{Main Components}
\textbf{Learnable Lookahead Tokens.}
\lookaheadkv \ performs KV cache eviction using a set of learnable special tokens during the prefill phase, followed by auto-regressive decoding with the preserved KV cache.
For a given input sequence \(X\), our framework appends a sequence of trainable soft lookahead tokens \(P = \{p_{1}, ..., p_{n_{\text{lookahead}}}\}\) whose queries in each attention head are used to estimate the attention pattern of the true model response.
In other words, these tokens are trained to compress the attention information of the true response to serve as the ``observation window'' in the eviction phase.
These are initialized randomly and added to the vocabulary before training.
Note that the lookahead tokens are used during the prefill stage only for eviction, and introduce no overhead for the decoding stage.

\textbf{Lookahead LoRA.}
To enhance the quality of estimation, we introduce lookahead LoRA, a novel low-rank adapter module that only activates for the lookahead tokens.
Lookahead LoRA provides complementary performance gains by allowing these tokens to learn richer representations, enabling their queries to more accurately predict token importance.
The selective activation mechanism of the LoRA modules ensures that the outputs of normal input tokens are unchanged, preserving the original model behavior.
Since the original model weights remain unaltered, \lookaheadkv \ modules can be selectively enabled or disabled depending on the particular requirements of a given application, thereby broadening the method’s applicability.
 
Combining the modules together, \lookaheadkv \ computes the queries and keys of the complete sequence as follows:
\begin{equation}
\mathbf{Q}_{\text{LKV}} = \begin{bmatrix} \mathbf{X} \\ \mathbf{P} \end{bmatrix} \mathbf{W}_{q} + \begin{bmatrix} \mathbf{0} \\ \mathbf{P} \end{bmatrix} \Delta \! \mathbf{W}_{q}
\;\;\;\;\;\;\;\;\;\;\;\;\;\;
\mathbf{K}_{\text{LKV}} = \begin{bmatrix} \mathbf{X} \\ \mathbf{P} \end{bmatrix} \mathbf{W}_{k} + \begin{bmatrix} \mathbf{0} \\ \mathbf{P} \end{bmatrix} \Delta \! \mathbf{W}_{k},
\end{equation}
where \(\mathbf{P} \in \mathbb{R}^{n_{\text{lookahead}} \times d}\) denotes the hidden states of the lookahead embeddings, and \(\Delta \! \mathbf{W}_{q}\), \(\Delta \! \mathbf{W}_{k}\) are the lookahead LoRA modules for query and key projections.
Similar to prior methods~\citep{li2024snapkv,cai2024pyramidkv,zhang2023h2o}, we use the attention matrix \({\mathbf{A}_{\text{LKV}}}=\operatorname{Softmax}(\frac{\, \mathbf{Q}_{\text{LKV}} \, \mathbf{K}_{\text{LKV}}^{\top} \,}{\sqrt{d}}),\) to estimate the importance score \(\tilde{s}_{j} = \frac{1}{n_{\text{lookahead}}} \sum_{i \,=\, n_{\text{in}} + 1}^{n_{\text{in}} + n_{\text{lookahead}}} \mathbf{A}_{\text{LKV}}{}_{\,i, j}\), and retain Top-K KV pairs with the highest importance scores.

\subsection{\lookaheadkv \ Training}
We train \lookaheadkv \ modules to compress the attention pattern of the true future response, using the model-generated responses as target.
Given a data pair \((X, Y)\), one iteration of \lookaheadkv \ training consists of the following steps:
\begin{enumerate}
  \item \textbf{GT Forward Pass.} For each layer \(l = 1, ..., L\) and head \(h = 1, ..., H\), the ground-truth importance scores \(\mathbf{s}_{\text{GT}}^{l,h}\) between the input prompt \(X\) and model-generated response \(Y\) are computed.
  \item \textbf{Lookahead Forward Pass.} For each layer \(l\) and head \(h\), we obtain the importance score estimates \(\mathbf{s}_{\text{LKV}}^{l,h}\) between the input prompt \(X\) and the lookahead tokens \(P\).
  \item \textbf{Loss Computation.} We first normalize all score vectors such that they sum to 1, and compute the average KL divergence loss between the GT and \lookaheadkv \ importance scores across all heads and layers:
  \begin{equation}
    \mathcal{L}_{\text{LKV}} \;=\; \frac{1}{L\cdot H} \sum_{l}^{L} \sum_{h}^{H} \operatorname{D_{KL}}\!\big( \, \hat{\mathbf{s}}_{\text{GT}}^{l, h} \,\; \Vert \;\, \hat{\mathbf{s}}_{\text{LKV}}^{l, h}\, \big).
  \end{equation}
    where \(\hat{\mathbf{s}}\) is the $L_1$-normalized importance scores such that \(\hat{\mathbf{s}} = \mathbf{s} / \|\mathbf{s}\|_1\). The loss is backpropagated to update the both the lookahead embeddings and LoRA modules, while all other LLM layers remain frozen, as shown in~\cref{fig1:main}.
    The pseudo-code for \lookaheadkv \ training and eviction is given in~\cref{pseudocode:training} and~\cref{pseudocode:eviction}.
\end{enumerate}

\textbf{Training Objective.}
Inspired from works on distilling attention scores~\citep{minilm,dpr}, we minimize the KL divergence between these normalized attention scores.
As our attentions scores are normalized, this KL divergence is equivalent to the popular ListNet~\citep{listnet} ranking loss, with $\phi$ of ListNet as identity instead of $\mathrm{exp}$.

\textbf{Lookahead LoRA Overhead.}
In principle, one can apply lookahead LoRA to any subset of the linear layers to tradeoff accuracy and latency.
However, even when lookahead LoRA is applied to every linear layer, there is a minor increase ($<$$1.3\%$) in latency compared to not using lookahead LoRA at all (see~\cref{tab:ablation} for ablation results), while significantly boosting performance compared to not using LoRA.
Consequently, we train \lookaheadkv \ with LoRA modules applied to all linear layers.

To avoid materializing the full attention score matrix, we use FlashAttention~\citep{dao2022flashattention} in the forward pass, coupled with eager attention for importance score computation and loss backpropagation, as detailed in~\cref{impl_opt}.

\section{Experiments}

\begin{figure}
    \centering
    \adjincludegraphics[width=\linewidth, max width=1.0\columnwidth]{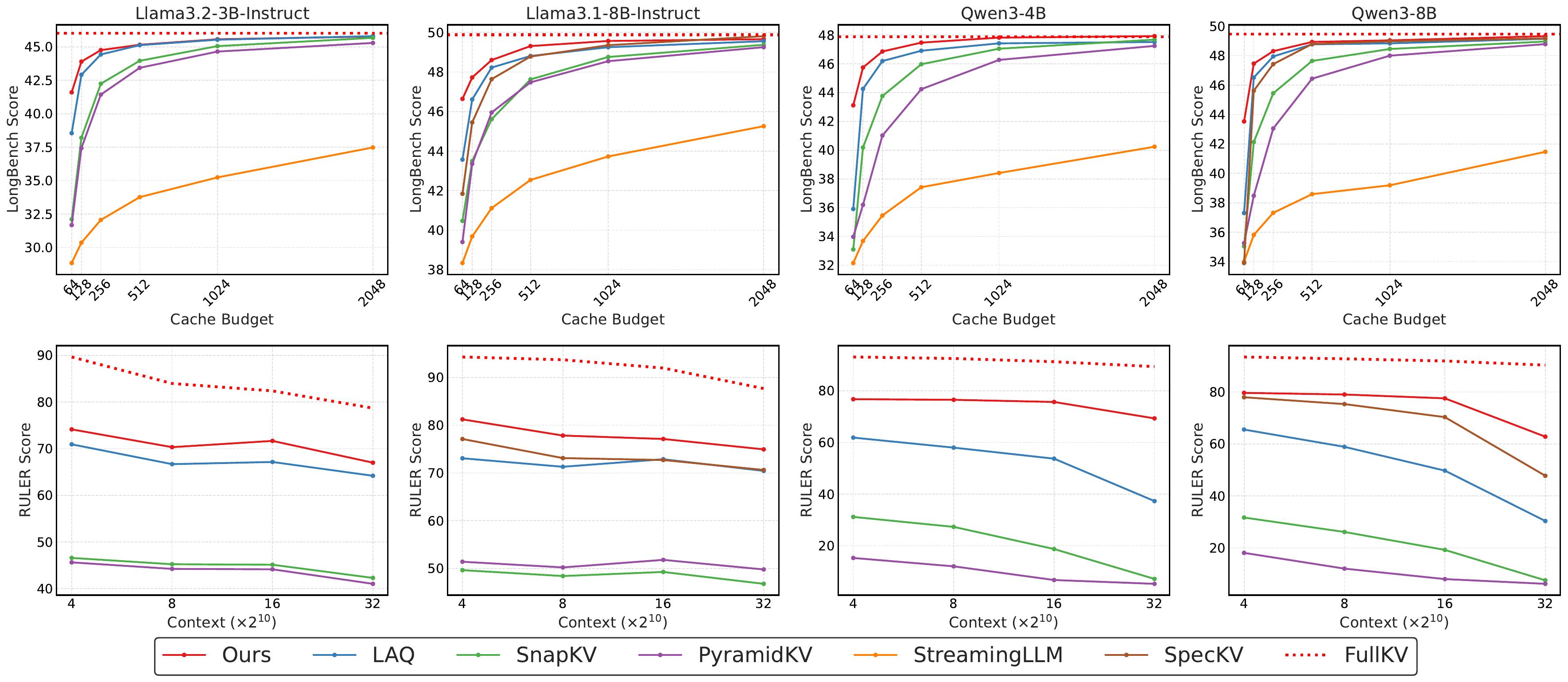}  
    \caption{
    Top row: Average LongBench results across multiple budgets and models. Bottom row: Average RULER results across varying context lengths with a fixed budget of $128$. Across all tested models, budgets and context lengths, \lookaheadkv \ consistently demonstrates superior performance.
    }\vspace{-0.3cm}
    \label{fig3:longbench_main}
\end{figure}

\subsection{Training}
\label{subsec:experiments/training}
\textbf{Dataset.}
To encourage the model to learn from diverse attention patterns, we curate training samples of varying lengths and sources, comprising both instruction-following datasets as well as pretraining texts.
We collect $50$K samples from the long\_sft subset of the ChatQA2~\citep{xu2024chatqa} dataset, $20$K samples from the Tulu~\citep{lambert_tulu_2025} instruction-following dataset, $7$K samples from the Stack~\citep{kocetkov2022stack}, and $9$K few-shot completion data samples that we create based on the training splits of the MetaMath, ARC, and HellaSwag datasets, originally curated in \citet{pal2024smaug}.
For instruction-following data, we remove the last assistant response and use the target model to obtain the \((X, Y)\) pairs of input prompt and model response.
For pretraining documents, we first truncate the text at random positions to obtain \(X\), and use the target model to complete the sequence to obtain \(Y\).
We limit the maximum input sequence length to $16$K, and generate all training responses using greedy decoding and max generation length of $512$.

\begin{minipage}[ht]{0.6\linewidth}
\textbf{Training Details.}
We apply \lookaheadkv \ on two widely used open-source architectures, LLaMA~\citep{dubey2024llama} and Qwen~\citep{yang2025qwen3technicalreport}, covering three model sizes each: LLaMA3.2-1B, LLaMA3.2-3B, LLaMA3.1-8B, Qwen3-1.7B, Qwen3-4B, and Qwen3-8B.
For all models, we set the lookahead size  \(n_{\text{lookahead}} = 32\), and apply LoRA to all projection and feed-forward modules (\(\mathbf{W}_{q}\), \(\mathbf{W}_{k}\), \(\mathbf{W}_{v}\), \(\mathbf{W}_{o}\), \(\mathbf{W}_{up}\), \(\mathbf{W}_{down}\), and \(\mathbf{W}_{gate}\)) with rank \(r = 8\) and scaling factor \(\alpha = 32\).
This configuration introduces less than $0.5$\% additional trainable parameters across all models, as summarized in~\cref{tab:trainable_params}.
Full hyperparameter settings are provided in~\cref{hparams_llama}.
\end{minipage}
\hfill
\begin{minipage}[ht]{0.37\linewidth}
\centering
\captionsetup{hypcap=false}\captionof{table}{Additional trainable parameters introduced by \lookaheadkv.}
\label{tab:trainable_params}
\resizebox{\textwidth}{!}{%
\begin{tabular}{l S[table-format=2.1M] S[table-format=1.2, round-mode=places, round-precision=2]}
    \toprule
    \multirow{2}{*}{Model} & \multicolumn{2}{c}{Trainable Params} \\
    \cmidrule(lr){2-3}
     & Params & \text{\% of Model} \\
    \midrule
    LLaMA3.2-1B   & 5.4\text{M}  & 0.44 \\
    LLaMA3.2-3B   & 11.9\text{M} & 0.37 \\
    LLaMA3.1-8B   & 20.6\text{M} & 0.26 \\
    \midrule
    Qwen3-1.7B    & 8.5\text{M} & 0.49 \\
    Qwen3-4B      & 16.2\text{M} & 0.40 \\
    Qwen3-8B      & 21.5\text{M} & 0.26 \\
    \bottomrule
\end{tabular}
}
\end{minipage}

\subsection{Evaluation Setup}
We evaluate our method on a comprehensive suite of benchmarks: LongBench~\citep{bai-etal-2024-longbench}, RULER~\citep{hsieh2024ruler}, LongProc~\citep{ye2025longproc}, and MT-Bench~\citep{zheng2023mtbench}.
LongBench is a multi-task benchmark that assesses the long-context understanding capability across diverse tasks, such as question answering, summarization, and code completion.
We report results on the $16$ English tasks, and use the average score as the main metric.
RULER is another multi-task synthetic benchmark, primarily comprising $13$ Needle-in-a-Haystack-style subtasks.
Each sample can be constructed at varying sequence lengths, allowing systematic evaluation of scaling behavior.
Similar to LongBench, we use average score as the main metric, and report the results at $4$K, $8$K, $16$K and $32$K context lengths.
We further evaluate the model's long-form output generation capability on the HTML to TSV task from LongProc, which involves converting structured information from long HTML documents into TSV format.
Finally, MT-bench provides a comprehensive multi-turn question set, spanning various domains such as writing, coding, and math.

\textbf{Baselines.}
We evaluate our method against popular KV cache eviction methods: \textbf{1) SnapKV}~\citep{li2024snapkv}, \textbf{2) PyramidKV}~\citep{cai2024pyramidkv}, and \textbf{3) StreamingLLM}~\citep{xiao2024streamingllm}.
We also compare our approach to stronger, more recent baselines that require costly approximate future response generation, such as \textbf{4) Lookahead Q-Cache} (LAQ)~\citep{wang2025laq}, and for 8B-scale models, \textbf{5) SpecKV}~\citep{galim2025speckv}.
In all experiments, Llama3.2-1B-Instruct and Qwen3-1.7B are used as draft models for Llama3.1-8B-Instruct and Qwen3-8B, respectively.
We follow the standard eviction configuration settings for all baseline methods, which we detail in \cref{app:hyperparams}.

\begin{table}[t]
\centering
\caption{MT‑Bench evaluation results. Bold and underlined values indicate best and second best scores, respectively.}
\label{tab:mtbench}
\resizebox{0.9\textwidth}{!}{%
\fontsize{10}{8}\selectfont
\begin{tabular}{l|r|cccccc}
\toprule
Model & Budget & PyramidKV & SnapKV & StreamingLLM & SpecKV & LAQ & \lookaheadkv\\
\midrule
& \multicolumn{7}{c}{\cellcolor{gray!25}\textit{FullKV score: 5.72}} \\
\cmidrule(lr){2-8}
\multirow{2}{*}{LLaMA3.2-1B} 
& 64   & 4.64 & 4.70 & 4.54 & N/A & \underline{5.03} & \textbf{5.21} \\
& 128  & 5.10 & 5.39 & 4.94 & N/A & \underline{5.45} & \textbf{5.60} \\
& 256  & 5.49 & \textbf{5.67} & 5.37 & N/A & \underline{5.64} & 5.62 \\
\midrule
& \multicolumn{7}{c}{\cellcolor{gray!25}\textit{FullKV score: 7.35}} \\
\cmidrule(lr){2-8}
\multirow{2}{*}{LLaMA3.2-3B} 
& 64   & 6.30 & 6.28 & 5.96 & \underline{6.52} & 6.48 & \textbf{6.87} \\
& 128  & 6.93 & \underline{7.03} & 6.42 & 7.02 & 6.93 & \textbf{7.26} \\
& 256  & 7.19 & \underline{7.30} & 7.20 & 7.28 & \textbf{7.43} & \underline{7.30} \\
\midrule
& \multicolumn{7}{c}{\cellcolor{gray!25}\textit{FullKV score: 7.77}} \\
\cmidrule(lr){2-8}
\multirow{2}{*}{LLaMA3.1-8B} 
& 64   & 6.85 & 6.80 & 6.17 & 6.77 & \underline{7.1} & \textbf{7.26} \\
& 128  & 7.39 & 7.50 & 6.84 & 7.34 & \underline{7.54} & \textbf{7.63} \\
& 256  & \underline{7.76} & 7.72 & 7.41 & 7.84 & 7.72 & \textbf{7.92} \\
\midrule

& \multicolumn{7}{c}{\cellcolor{gray!25}\textit{FullKV score: 7.19}} \\
\cmidrule(lr){2-8}
\multirow{2}{*}{Qwen3-1.7B} 
& 64   & 5.81 & 5.95 & 5.83 & N/A & \underline{6.19} & \textbf{6.70} \\
& 128  & 6.38 & 6.65 & 6.16 & N/A & \underline{6.91} & \textbf{7.12} \\
& 256  & 6.90 & 6.94 & 6.91 & N/A & \underline{7.02} & \textbf{7.20} \\
\midrule
& \multicolumn{7}{c}{\cellcolor{gray!25}\textit{FullKV score: 8.02}} \\
\cmidrule(lr){2-8}
\multirow{2}{*}{Qwen3-4B} 
& 64   & 6.85 & 6.60 & 6.24 & 7.05 & 7.06 & \textbf{7.69} \\
& 128  & 7.55 & 7.71 & 7.24 & 7.78 & 7.70 & \textbf{8.12} \\
& 256  & 7.90 & \textbf{8.20} & 7.87 & 8.11 & \underline{8.12} & 8.06 \\
\midrule
& \multicolumn{7}{c}{\cellcolor{gray!25}\textit{FullKV score: 8.48}} \\
\cmidrule(lr){2-8}
\multirow{2}{*}{Qwen3-8B} 
& 64   & 7.33 & 7.26 & 6.81 & \underline{7.69} & 7.58 & \textbf{8.04} \\
& 128  & 7.85 & 7.94 & 7.64 & 7.97 & \underline{8.24} & \textbf{8.41} \\
& 256  & 8.42 & 8.43 & 8.34 & 8.45 & \textbf{8.56} & \underline{8.51} \\
\bottomrule
\end{tabular}%
}
\end{table}

\subsection{Performance Results}
\label{subsec:experiments/longbench}
\textbf{LongBench Evaluation.}
\cref{fig3:longbench_main} shows the average LongBench scores of \lookaheadkv \ and baselines, across cache budget settings ranging from $64$ to $2048$.
Our method consistently demonstrates superior performance across all models and all budgets tested, demonstrating the effectiveness and robustness of our approach.
Overall, results show that expensive draft-based methods (e.g., LAQ and SpecKV) outperform simple baselines, corroborating that employing approximate future response for importance estimation is effective.
Nevertheless, our method significantly outperforms draft-based approaches, especially at lower budget settings, highlighting that learning to estimate future importance is crucial for performance preservation.
Due to space limitations, we report the results of 1B-scale models in~\cref{app:additional_experiments}.

\textbf{RULER Evaluation.}
We report the RULER evaluation results of all methods with a fixed budget of $128$ in~\cref{fig3:longbench_main} (1B-scale results are provided in \cref{app:additional_experiments}).
\lookaheadkv \ consistently outperforms other baseline approaches here as well, maintaining strong performance across all evaluated context lengths.
Despite being trained on a maximum sequence length of $16$K, \lookaheadkv \ effectively generalizes to a longer context length of $32$K.
We conduct additional experiments on the impact of training context length in~\cref{subsec:analysis/shorter_context}.

\textbf{Long-form Output Evaluation.}
To further validate \lookaheadkv's ability to generate long-form outputs, we evaluate our method on the HTML to TSV task from LongProc.
We assess \lookaheadkv \ and baseline methods under two input–output settings: $12$K–$0.5$K and $23$K–$2$K tokens, both at a fixed cache budget ratio of $30$\%.

\begin{wrapfigure}[16]{r}{0.48\textwidth}
\setlength{\intextsep}{0pt}
  \centering
    \includegraphics[width=0.48\textwidth,trim={0cm 0cm 3cm 0cm},clip]{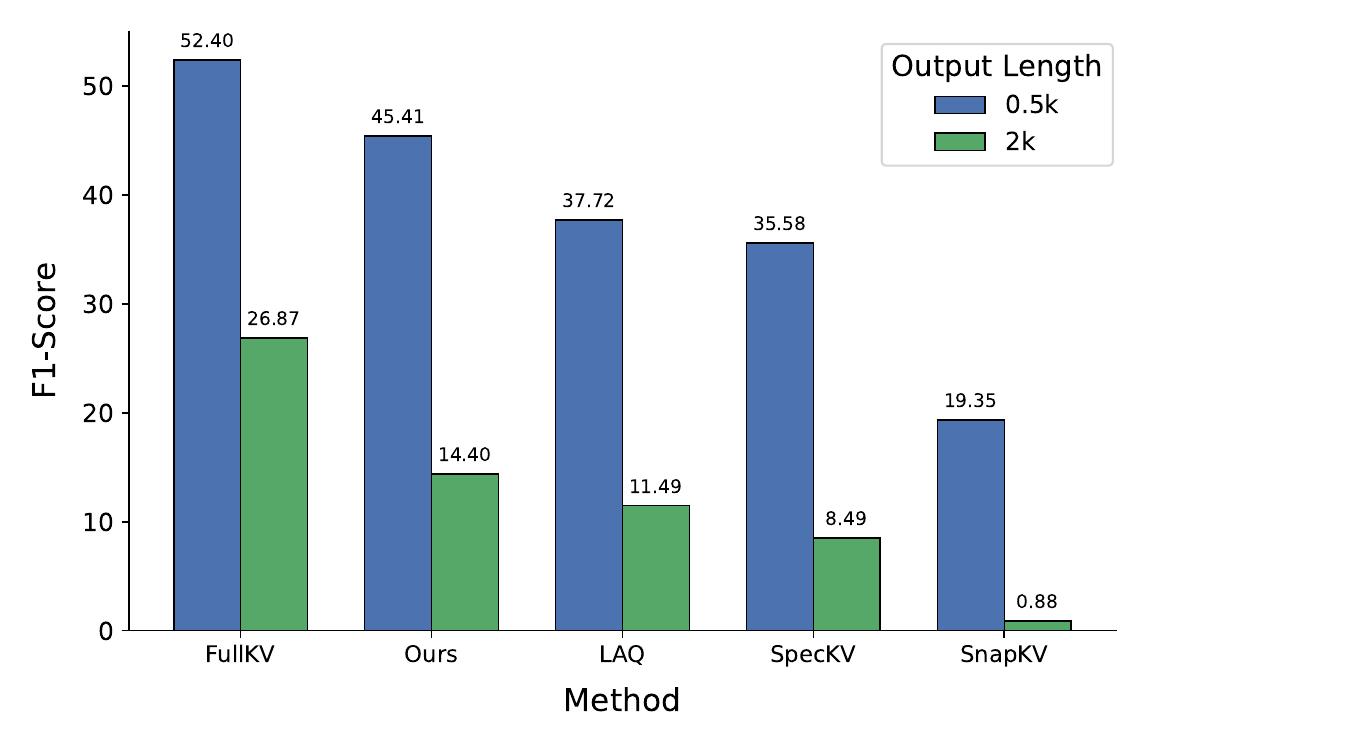}
  \caption{HTML-to-TSV evaluation results using LLaMA3.1-8B.}
  \label{fig5:html}
\end{wrapfigure}

\cref{fig5:html} presents the results on the HTML to TSV task using LLaMA-3.1-8B.
Across both sequence-length configurations, \lookaheadkv \ consistently outperforms prior approaches.
We hypothesize that \lookaheadkv, learning to predict the attention pattern of the entire future response, is particularly superior in long-form generation tasks compared to draft-based methods that rely only on partial future response as the observation window.

\textbf{Multi-turn Evaluation.}
To test our method under multi-turn conversation setting, we evaluate \lookaheadkv \ and baselines on MT-Bench~\citep{zheng2023mtbench}.
The generated responses are evaluated using Qwen3-235B-A22B as the LLM judge.
The results in~\cref{tab:mtbench} indicate that \lookaheadkv \ is either on par or superior across all models and budgets tested.
\lookaheadkv \ is particularly robust in lower budget settings (e.g., \(C = [64, 128]\)), where it consistently outperforms all other methods.
\section{Analysis}

\subsection{Efficiency Comparison}

\begin{table}[t]
  \centering
  \caption{Theoretical and empirical cost analysis of LLaMA3.1-8B at $C=128$.}
  \label{tab:kv-theoretical-practical}
  \begingroup
    \setlength{\tabcolsep}{4pt}      
    \renewcommand{\arraystretch}{1.35} 

    \resizebox{\textwidth}{!}{%
    \fontsize{7}{6}\selectfont
      \begin{tabular}{c l S[table-format=4, round-mode=places, round-precision=0] S[table-format=3, round-mode=places, round-precision=0] S[table-format=4, round-mode=places, round-precision=0] S[table-format=3.2, round-mode=places, round-precision=2] S[table-format=4, round-mode=places, round-precision=0] S[table-format=3, round-mode=places, round-precision=0]}
        \toprule
        & & \multicolumn{4}{c}{\textbf{Theoretical Cost}} & \multicolumn{2}{c}{\textbf{Empirical Cost}} \\
        \cmidrule(lr){3-6}\cmidrule(lr){7-8}
        \shortstack[c]{\textbf{Context}\\\textbf{Length}} &
        \textbf{Method} &
        \shortstack[c]{\textbf{Compute}\\\textbf{(TFLOPs)}} &
        \shortstack[c]{\textbf{Memory Traffic}\\\textbf{(GB)}} &
        \shortstack[c]{\textbf{TTFT}\\\textbf{(ms)}} &
        \shortstack[c]{\textbf{TTFT}\\\textbf{Overhead (ms)}} &
        \shortstack[c]{\textbf{TTFT}\\\textbf{(ms)}} &
        \shortstack[c]{\textbf{TTFT}\\\textbf{Overhead (ms)}} \\
        \midrule
        \multirow{5}{*}{$8$K}
          & Forward Pass Only    & 136.00  & 13.00  & 256.99 & N/A      & 290.81  & N/A      \\
          & \lookaheadkv         & 136.53  & 13.05  & 258.03 & 1.0345   & 301.69 & 10.8800  \\
          & SnapKV               & 136.00  & 13.02  & 257.00 & 0.0085   & 310.98 & 20.1700  \\
          & SpecKV               & 159.12  & 81.07  & 336.52 & 79.5264  & 411.32 & 120.5100 \\
          & LAQ                  & 137.06  & 444.52 & 491.58 & 234.5877 & 800.19 & 509.3800 \\
        \midrule
        \multirow{5}{*}{$32$K}
          & Forward Pass Only    & 928.00  & 13.00  & 1753.59& N/A      & 1760.22  & N/A      \\
          & \lookaheadkv        & 928.91  & 13.05  & 1755.33& 1.7431   & 1798.26 & 38.0400   \\
          & SnapKV               & 928.00  & 13.02  & 1753.60& 0.0085   & 1837.89 & 77.6700  \\
          & SpecKV               & 1115.09 & 105.82 & 2156.39& 402.7969 & 2263.09 & 502.8700 \\
          & LAQ                  & 929.81  & 450.52 & 1992.85& 239.2601 & 2313.90 & 553.6800 \\
        \bottomrule
      \end{tabular}%
    }
  \endgroup
  \label{table:cost}
\end{table}
Efficiency is assessed by measuring the Time-To-First-Token (TTFT) for LLaMA-3.1-8B across multiple context lengths. While we utilize official codebases for most baselines, LAQ was re-implemented due to the absence of an official release.
Because the latency of a method can vary significantly depending on the implementation, we conduct rigorous analysis and derive the theoretical latency for each method, based on the analytical model proposed in~\citet{davies2025efficientllminferencebandwidth}.
We discuss further details in~\cref{app:cost-details}.

\cref{table:cost} presents the results of the TTFT analysis for $8$K and $32$K context lengths (see \cref{tab:kv-theoretical-practical-full} for $4$K and $16$K results).
Overall, we observe that draft-based methods incur significant overhead, either due to increased computation (SpecKV) or memory traffic (LAQ).
On the contrary, \lookaheadkv \ requires marginal additional cost across all tested context lengths, reducing eviction overhead by $14.5$× compared to LAQ at $32$K sequence length.

\subsection{Effect of Stochastic Decoding}
We analyze the effect of stochastic generation on \lookaheadkv's performance by evaluating our method using two temperature settings: \([0.2, 0.8]\).
Results in~\cref{tab:stochastic} show that \lookaheadkv \ maintains superior performance over all other baselines across all temperature settings.
Further, we observe that performance degradation at high temperature setting (3-4\% at \(T=0.8\)) is consistent across all methods, including FullKV, indicating that stochasticity in inference affects all approaches similarly.
We further discuss the interplay between stochastic decoding for training data generation and \lookaheadkv \ performance in~\cref{app:stochastic}
\begin{table}[t]
    \centering
    \caption{Average LongBench performance at different temperature settings on LLaMA3.1-8B, with $C = 128$. \lookaheadkv \ outperforms baselines across all tested temperature settings.}
    \resizebox{\linewidth}{!}{
    \fontsize{7}{7}\selectfont
    \begin{tabular}{l|c|cccc}
        \toprule
        Method & FullKV & SnapKV & SpecKV & LAQ & \lookaheadkv \\
        \midrule
        Greedy      & 49.88 & 43.50 & 45.45 & 46.61 & \textbf{47.72} \\
        \midrule
        $T=0.2$   & 49.58 \scriptsize{(-0.60\%)} & 43.29 \scriptsize{(-0.48\%)} & 44.99 \scriptsize{(-1.01\%)} & 46.73 \scriptsize{(+0.26\%)} & \textbf{47.75} \scriptsize{(+0.06\%)} \\
        $T=0.8$   & 47.82 \scriptsize{(-4.13\%)} & 41.39 \scriptsize{(-4.85\%)} & 43.43 \scriptsize{(-4.44\%)} & 45.27 \scriptsize{(-2.87\%)} & \textbf{45.81} \scriptsize{(-4.00\%)} \\
        \bottomrule    
    \end{tabular}
    }
    \label{tab:stochastic}    
\end{table}

\subsection{Ablation on Trainable Modules}
We study the impact of lookahead size \(n_{\text{lookahead}}\) and LoRA placement through a 2D ablation across six lookahead sizes ($4$, $8$, $16$, $32$, $64$, $128$) and three configurations: \texttt{emb-only} (No LoRA applied), \texttt{QV} (LoRA applied to Q and V), and \texttt{all} (LoRA applied to all linear layers).
The results indicate that both larger lookahead windows and broader LoRA coverage generally improve average LongBench performance.
However, performance gains saturate at \(n_{\text{lookahead}} = 32\); increasing the lookahead size beyond this point yields diminishing returns while incurring a noticeable increase in inference overhead.
On the other hand, applying lookahead LoRA to all layers results in relatively minor rise in TTFT while significantly improving the performance across all lookahead sizes.
Based on this analysis, we set \(n_{\text{lookahead}} = 32\) and apply LoRA to all linear modules in our main experiments.

\begin{table}[t]
    \centering
    \caption{2D ablation across lookahead sizes and trainable modules, on LLaMA3.2-1B. Average LongBench scores with cache budget of 64 and TTFT overhead are reported.}  
    \resizebox{\textwidth}{!}{%
    \fontsize{16}{18}\selectfont
        \begin{tabular}{
            l |
            S[table-format=2.1, round-mode=places, round-precision=1]S[table-format=1.1, round-mode=places, round-precision=1]
            S[table-format=2.1, round-mode=places, round-precision=1]S[table-format=1.1, round-mode=places, round-precision=1]
            S[table-format=2.1, round-mode=places, round-precision=1]S[table-format=1.1, round-mode=places, round-precision=1]
            S[table-format=2.1, round-mode=places, round-precision=1]S[table-format=1.1, round-mode=places, round-precision=1]
            S[table-format=2.1, round-mode=places, round-precision=1]S[table-format=1.1, round-mode=places, round-precision=1]
            S[table-format=2.1, round-mode=places, round-precision=1]S[table-format=1.1, round-mode=places, round-precision=1]
        }  
            \toprule  
            & \multicolumn{2}{c}{\(n_{\text{lookahead}}=4\)} &  
              \multicolumn{2}{c}{\(n_{\text{lookahead}}=8\)} &  
              \multicolumn{2}{c}{\(n_{\text{lookahead}}=16\)} &  
              \multicolumn{2}{c}{\(n_{\text{lookahead}}=32\)} &
              \multicolumn{2}{c}{\(n_{\text{lookahead}}=64\)} &
              \multicolumn{2}{c}{\(n_{\text{lookahead}}=128\)}\\   
            \cmidrule(lr){2-3} \cmidrule(lr){4-5}  
            \cmidrule(lr){6-7} \cmidrule(lr){8-9}
            \cmidrule(lr){10-11} \cmidrule(lr){12-13}
            Module &  \text{score} & \text{overhead(\%)} &  
                      \text{score} & \text{overhead(\%)} &  
                      \text{score} & \text{overhead(\%)} &
                      \text{score} & \text{overhead(\%)} &
                      \text{score} & \text{overhead(\%)} &
                      \text{score} & \text{overhead(\%)} \\   
            \midrule  
            \texttt{emb-only} &  
              25.53 & 3.36 &  
              25.67 & 3.77 &  
              26.36 & 3.42 &  
              26.40 & 4.18 &
              25.8 & 7.3  &
              26.2 & 10.7 \\[2mm]  
            \texttt{QV} &  
              26.45 & 3.65 &  
              26.40 & 4.08 &  
              26.87 & 3.99 &  
              26.85 & 4.42 &
              26.7 & 7.7 &
              27.1 & 10.7 \\[2mm]  
            \texttt{all} &  
              26.60 & 4.20 &  
              27.02 & 4.20 &  
              27.01 & 4.67 &  
              27.12 & 4.96 &
              27.1 & 8.5 & 27.0 & 11.0 \\
            \bottomrule  
        \end{tabular}
    }
    \label{tab:ablation}  
\end{table}

\subsection{Robustness to Training Context Length}
\label{subsec:analysis/shorter_context}
\begin{minipage}[ht]{0.54\linewidth}
Transformer-based language models trained with fixed context lengths often struggle to generalize beyond their training window.
Similarly, one may raise concern about the context length generalization of our method.
To examine this effect, we apply \lookaheadkv \ training to LLaMA-3B with limited training context lengths of $2$K, $4$K, and $8$K, and evaluate on RULER (\cref{fig:rule_shorter_context}).
We observe that while longer training context length yields better performance as expected, training on shorter contexts still remains effective with relatively minor degradation in performance, demonstrating that our method generalizes robustly to unseen sequence lengths.
\end{minipage}
\hfill
\begin{minipage}[ht]{0.42\linewidth}
    \includegraphics[width=0.98\linewidth]{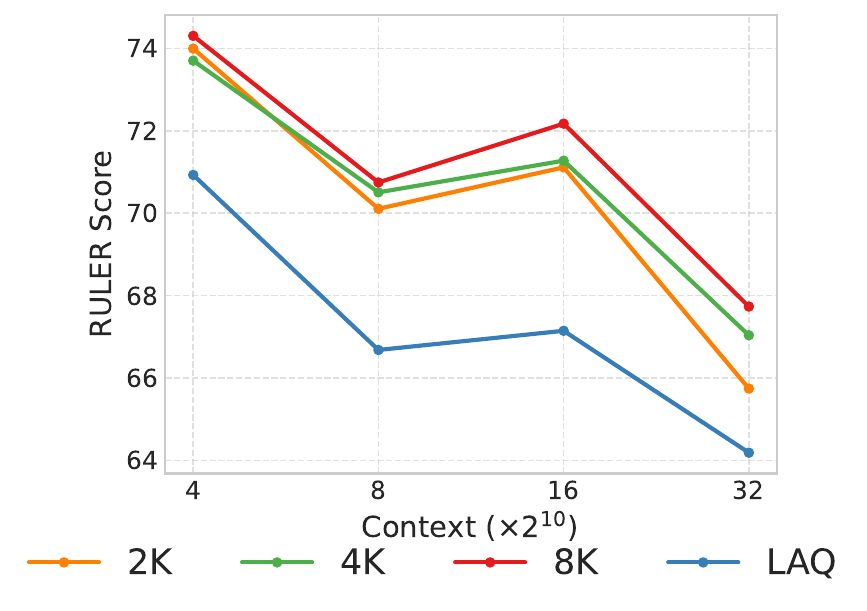}  
    \captionsetup{hypcap=false} \captionof{figure}{RULER evaluation on \lookaheadkv \ trained with shorter contexts.}
    \vspace{-0.3cm}
    \label{fig:rule_shorter_context}
\end{minipage}
\section{Related Work}\label{sec:related_work}
   
\textbf{KV Cache Eviction.}
Early analyses revealed that attention scores tend to be sparse~\citep{zhang2023h2o}, implying that only a small subset of KV entries substantially contributes to the attention output.
Further, subsequent work showed that the importance of these tokens remains stable throughout generation, i.e., tokens deemed important early on tend to stay important~\citep{liu2023scissorhands}.
These observations motivated a range of eviction methods aimed at discarding unimportant KV entries while preserving model performance.
Representative methods such as H2O~\citep{zhang2023h2o}, NACL~\citep{chen2024nacl} and TOVA~\citep{oren-etal-2024-transformers} rely on attention scores to estimate token importance and evict low-importance KV pairs. In contrast, other works propose to use alternative importance metrics for eviction~\citep{park2025keydiff, guo-etal-2024-vam, geng2025andpro} to improve importance estimation or address the challenge of materializing the full attention matrix.

\textbf{Prefill KV Cache Eviction.}
A specific line of work, which we discuss extensively in our paper, focuses on eviction of prefill KV cache.
SnapKV~\citep{li2024snapkv} introduced the notion of an ``observation window'' consisting of the suffix of the input prompt, which is used to predict important tokens to keep for subsequent response generation.
SpecKV~\citep{galim2025speckv} proposed to generate an approximate response with a smaller model and use the resulting tokens as a more reliable observation window for future importance prediction.
Lookahead Q-Cache~\citep{wang2025laq} first applies a simple eviction method, such as SnapKV, to obtain a partial low‑cost draft response, then re‑evicts KV entries based on the importance scores derived from the draft.
KVzip~\citep{kim2025kvzip} adopts a query‑agnostic strategy by inserting a repeated prompt and measuring which KV entries are essential for accurately reconstructing the input.
Orthogonal to these approaches, several works proposed to allocate non-uniform budgets for each layer~\citep{cai2024pyramidkv} and head~\citep{feng2024ada} to further improve performance.

\textbf{Prompt Tuning for Task Adaptation.}
Another line of work closely related to ours is parameter-efficient finetuning through learned prompts.
Prompt Tuning~\citep{lester2021prompttuning} inserts a sequence of continuous, learnable embeddings into the frozen LLM for downstream task adaptation, while Prefix‑Tuning~\citep{li2021prefix} extends this idea by pre-pending learned vectors across multiple layers.
Further, P‑Tuning v2~\citep{liu2021ptuning2} demonstrated that prompt‑based adaptation scales well across a wide range of model sizes.
Unlike conventional prompt‑tuning methods that aim to improve task performance, our work leverages learned prompts to predict internal model statistics, thereby enhancing computational efficiency rather than accuracy.

Training objectives similar to ours have been used in distillation~\citep{minilm}, or in ranking/retrieval~\citep{listnet,dpr}.
Some contemporaneous works~\citep{greenewald2025activated,peng2025tara,samragh2025llm} also propose LoRA modules that selectively activate only for some tokens.

\section{Conclusion and Limitation}
We introduce \lookaheadkv, a trainable KV cache eviction framework that accurately predicts token importance without relying on explicit draft generation.
The method augments a frozen LLM with a small set of learnable lookahead tokens and lookahead LoRA modules that activate only on these tokens.
Trained to match ground-truth importance distributions across layers and heads, \lookaheadkv \ achieves performance superior to costly draft-based approaches.
Across a wide range of model families and long-context benchmarks, our approach consistently outperforms prior methods, especially in low-budget regimes. It introduces less than 0.5\% additional parameters and incurs only a marginal increase in prefill latency.

Due to limited compute resources, we were unable to conduct experiments on larger-sized models.
However, experimental results indicate that \lookaheadkv \ improves both performance and latency of KV cache eviction across a variety of model sizes.
\lookaheadkv \ currently focuses on the prefill KV cache eviction; extending \lookaheadkv \ to also perform decoding-stage eviction remains an interesting future work.

\clearpage
\section*{Acknowledgment}
We would like to thank Daehyun Kim, Ph.D., Hyeonmok Ko, Ph.D., Ho-young Kim, Anshumann, Mohd Abbas Zaidi, and Harshith Goka for their helpful discussions and generous support in this work.
\bibliography{iclr2026_conference}
\bibliographystyle{iclr2026_conference}

\clearpage
\appendix
\section{Pseudo-code}
\label{app:pseudocode}
The pseudocode for \lookaheadkv \ training and eviction is described in~\cref{pseudocode:training} and~\cref{pseudocode:eviction}, respectively.
\begin{algorithm}
\caption{\lookaheadkv \ Training}\label{pseudocode:training}
\begin{algorithmic}[1]  
\Require dataset $\mathcal{D}$ of input-response pairs
\State $\texttt{scores} \gets []$                \Comment{GT importance scores}
\State $\texttt{estimates} \gets []$             \Comment{score estimates using \lookaheadkv}
\For{each training sample $(X,Y)$ in dataset $D$}  
    \For{each layer $l$}                            \Comment{GT pass}
        \For{each head $h$ in layer $l$}  
            \State $S \gets \text{GT importance score for head }(l,h)$  
            \State $\texttt{scores}.\texttt{append}(S)$  
        \EndFor  
    \EndFor
    
    \For{each layer $l$}                            \Comment{lookahead pass}
        \For{each head $h$ in layer $l$}  
            \State $\hat S \gets \text{importance scores using lookahead embeddings for head }(l,h)$  
            \State $\texttt{estimates}.\texttt{append}(\hat S)$  
        \EndFor  
    \EndFor  

    \State $L \gets 0$                         \Comment{compute loss}  
    \ForAll{$(S,\hat S)$ in $\texttt{scores}$, $\texttt{estimates}$}  
        \State $L \gets L + \operatorname{D_{KL}}\bigl(\frac{S}{\|S\|_1} \,\|\, \frac{\hat{S}}{\|\hat{S}\|_1}\bigr)$  
    \EndFor  
    \State $L \gets \frac{L}{|\texttt{scores}|}$
    \State $L.\text{backward}()$
\EndFor
\end{algorithmic}
\end{algorithm}
\begin{algorithm}  
\caption{\lookaheadkv \ Eviction}\label{pseudocode:eviction}  
\begin{algorithmic}[1]  
\Require Input prompt $X=\{x_1,\dots,x_{n_{\text{in}}}\}$
\Require cache budget $k$
\State Append learned lookahead tokens to input and compute the sequence embeddings $\hat{\mathbf{X}} = [\mathbf{X} \; \mathbf{P}]^{\top}$ \Comment{shape: $(n_{\text{in}} + n_{\text{lookahead}}) \; \times \; d$}
\State Perform a prefill forward pass with $\hat{\mathbf{X}}$: 

\For{each layer $l$}  
    \For{each head $h$}
        \State $\mathbf{A} \gets
               \operatorname{Softmax}\!\Big(  
               \frac{QK^{\!\top}}  
                    {\sqrt{d}}\Big)$ \Comment{shape: $(n_{\text{in}} + n_{\text{lookahead}}) \; \times \; (n_{\text{in}} + n_{\text{lookahead}})$}
        \State $\hat{\mathbf{A}} \gets \mathbf{A} [n_{\text{in}}: \;,\; :n_{\text{in}}]$ \Comment{attention between lookahead tokens and input prompt}

        \State $\mathbf{s} \gets  \operatorname{MeanReduce}(\hat{\mathbf{A}})$ 

        \State $\mathbf{s} \gets  \operatorname{Pooling}(\mathbf{s})$ \Comment{score vector, shape: $1 \;\times\; n_{\text{in}}$}  
        
        \State $\mathcal{I} \gets \text{TopK}(\mathbf{s},\,k)$

        \State $K^{\text{kept}} \gets K[\mathcal{I}]$  
        \State $V^{\text{kept}} \gets V[\,\mathcal{I}]$  
        \State Cache $(K^{\text{kept}},\,V^{\text{kept}})$  \Comment{evict unimportant KV pairs}

        \State Compute attention output for MLP layer
    \EndFor
    \State Compute MLP output for next layer
\EndFor  

\State \Return
\end{algorithmic}  
\end{algorithm}  

\section{Theoretical Estimation Details}\label{app:cost-details}
This section details our methodology for the theoretical estimation of the Time-to-First-Token (TTFT) latency for various KV cache eviction algorithms. Our analysis is based on the analytical model for FLOPs and memory traffic proposed by \citet{davies2025efficientllminferencebandwidth}.
To align configurations of theoretical estimates with those of actual measurements, we simulate the execution of LLaMA3.1-8B on a single NVIDIA H100 80GB GPU with a batch size of 1, assuming all weights and activations are in half-precision. We set KV cache budget size of 128, lookahead size as 32, and window size as 32. We only consider tensor operations which are dominant parts of the computations. To provide estimates that closely reflect real-world performance, our calculations incorporate practical hardware utilization by assuming a flops efficiency of 0.7 and a memory efficiency of 0.9, as described in \citet{llm-analysis-chengli}.

To isolate the specific overhead introduced by each eviction algorithm, we first establish a baseline by calculating the theoretical latency of a single forward pass. The TTFT overhead for each eviction method is then determined by subtracting this baseline forward pass latency from the method's total estimated TTFT. For LAQ, the total latency is calculated by summing the costs of its three consecutive steps—the first eviction, low-cost generation of pseudo response, and the second eviction. Similarly, the total latency of SpecKV is estimated by aggregating the latencies of its draft prefill, draft decode, and target model eviction phases. A comprehensive implementation of the code to derive theoretical estimates of all baselines is available in the Supplementary Materials.

\section{Implementation Optimization}
\label{impl_opt}
Efficient attention implementations such as FlashAttention~\citep{dao2022flashattention} do not materialize the full attention score matrix, but is required in our setting to compute importance scores and enable gradient backpropagation. A possible solution is to compute the complete attention matrix using native PyTorch (i.e., eager attention), but this quickly leads to an out-of-memory error as the matrix size grows quadratically with the sequence length, which is incompatible with our training setting (up to $16$K sequence length). Fortunately, for our objective, we only require the cross-attention scores between the generated response and the entire input sequence, and the response length is typically much shorter than the input prompt. 

Leveraging this observation, we adopt the following approach: for the attention layer's forward pass, we use flash attention, while for importance score computation and backpropagation, we employ eager attention.
This reduces the memory requirement of eager attention from $\mathcal{O}((|X|+|Y|)^2)$ to $\mathcal{O}(|X|\cdot|Y|+|Y|^2)$, where $|X|$ and $|Y|$ denote the lengths of the input prompt and model response, respectively, with $|X| \gg |Y|$.

\newpage

\section{Need for Data Generation}
One of the requirements of \lookaheadkv \ training is that the target model's generated responses must be available as training data. However, generating these responses from the model can sometimes be costly, e.g., when applying \lookaheadkv \ across multiple models. Hence, to assess whether this requirement of can be relaxed, we evaluate an alternative setting where training uses the responses from the source datasets instead of model-generated outputs.

\begin{figure}[h]
    \centering
    \adjincludegraphics[width=0.5\linewidth, max width=0.5\columnwidth]{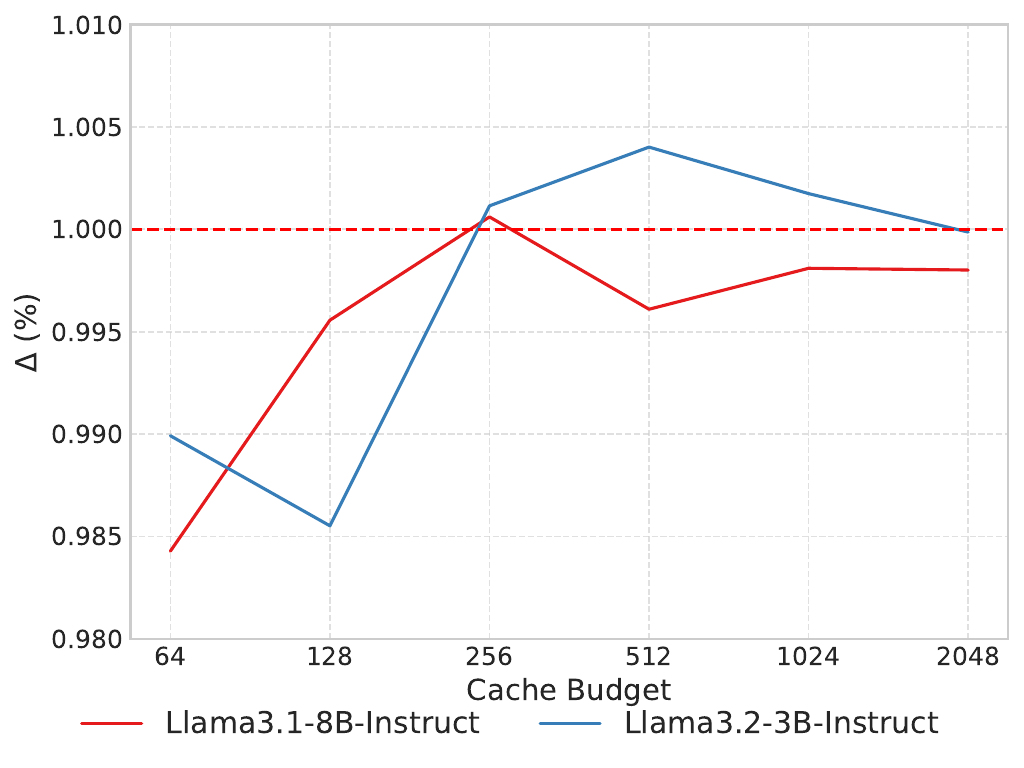}  
    \caption{
    Performance ratio of training using model-generated data vs. source data.
    }\vspace{-0.3cm}
    \label{fig:gt}
\end{figure}

We observe in~\cref{fig:gt} that this substitution leads to a relatively minor drop in average LongBench performance in lower-budget regimes. We hypothesize that if the attention distribution of the model-generated responses and that of the source dataset responses are moderately similar, our method can still successfully learn to accurately predict the importance scores. Overall, these results suggest that, in scenarios where training data generation is impractical, using source responses provides a viable and effective alternative.

\newpage

\section{Additional Results}
\label{app:additional_experiments}
In this section, we provide additional experimental results excluded from the main text due to page limitations.

\subsection{RULER Evaluation on Longer Contexts}
To explore the capability of \lookaheadkv \ on longer contexts, we evaluate our method on RULER at 64K and 128K context lengths using LLaMA3.1-8B-Instruct with a cache budget of 128. We randomly sample 50 examples per task from the RULER benchmark. As shown in~\cref{tab:ruler-longer-context}, \lookaheadkv \ achieves the best performance at these context lengths as well, showing that the effectiveness of our method scales to even longer context lengths.
\begin{table}[ht]  
    \centering  
    \caption{RULER evaluation results on longer context lengths using Llama3.1-8B-Instruct at $C=128$.}
    \label{tab:ruler-longer-context}  
    \begin{tabular}{l|S[table-format=2.2]S[table-format=2.2]S[table-format=2.2]S[table-format=2.2]S[table-format=2.2]}
        \toprule
        {Context Length} & {FullKV} & {\lookaheadkv} & {SnapKV} & {SpecKV} & {LAQ} \\
        \midrule
        64K  & 84.15 & \textbf{69.45}   & 39.64 & 64.02 & 64.10 \\
        128K & 73.72 & \textbf{54.83}   & 30.56 & 52.62 & 50.67 \\
        \bottomrule
    \end{tabular}
\end{table}

\subsection{Effect of combining suffix window}
To test the effect of incorporating suffix window, as proposed in SnapKV~\citep{li2024snapkv}, we augment \lookaheadkv \ by also including queries of the last 32 prompt tokens for importance score estimation. As shown in~\cref{tab:suffix_window}, we observe a slight drop in performance when SnapKV importance scores are included. The degraded performance when averaging \lookaheadkv \ importance scores with SnapKV scores, compared to using \lookaheadkv \ scores alone, indicates that the importance predicted by our method is superior to SnapKV.
\begin{table}[htbp]
\caption{Average LongBench scores using \lookaheadkv \ window only and \lookaheadkv \ + SnapKV-style suffix window, evaluated using LLaMA3.2-1B-Instruct with $C = 64$.}
\centering
\resizebox{0.5\linewidth}{!}{
\begin{tabular}{c|cc}
\toprule
FullKV & \lookaheadkv \ & \lookaheadkv \ \\
& & + Suffix Window \\
\midrule
32.01 & \textbf{29.10} & 28.52 \tiny{(-1.99\%)} \\
\bottomrule
\end{tabular}
}
\label{tab:suffix_window}
\end{table}

\subsection{Discussion of Generation Stochasticity in \lookaheadkv \ Training}
\label{app:stochastic}
For \lookaheadkv \ training, various stochastic decoding methods may be employed to generate training data. One may hypothesize that the attention matrices induced by responses generated with higher stochasticity may diverge significantly from those from greedy responses, potentially limiting the generalizability of \lookaheadkv \ modules trained exclusively on greedy responses to stochastic inference scenarios. To investigate this, we quantify the similarity between importance score vectors induced by greedy responses and those generated under varying temperature settings.

\cref{tab:similarity} presents recall@512 and Kendall rank correlation coefficients comparing importance scores induced by greedy decoding against stochastic decoding at multiple temperatures using LLaMA3.1-8B. The scores are averaged over 30 randomly selected samples from our training data, across all layers and heads. Even at relatively high temperature ($T= 0.8$), we observe strong persistence of attention patterns. Notably, the deviation is smaller than that induced by responses of a speculative model (Llama3.2-1B, equivalent to the SpecKV setting). This indicates that the ground-truth importance scores derived from stochastically generated responses are highly similar to those from greedy responses, which in turn indicates that greedy-generated training data provides sufficiently robust learning signals for stochastic settings.
\begin{table}[h]  
    \centering  
    \caption{Importance score similarity with stochastic response using various temperatures vs. greedy response on LLaMA3.1-8B. LLaMA3.2-1B presents the similarity of importance scores using greedy response generated with LLaMA3.2-1B vs. LLaMA3.1-8B.}
    \begin{tabular}{l|
    S[table-format=2.2]
    S[table-format=2.2] 
    S[table-format=2.2]
    S[table-format=2.2]
    S[table-format=2.2]
   }  
    \toprule  
    Generation Method &  {$T = 0.2$} &  {$T = 0.4$} &  {$T = 0.6$} &  {$T = 0.8$} &  {LLaMA3.2-1B} \\  
    \midrule  
    {Recall@512 (\%)} & 95.06 & 93.73 & 91.40 & 91.37 & 88.66 \\  
    {Kendall's Tau}       & 91.44 & 88.63 & 84.61 & 84.79 & 80.05 \\  
    \bottomrule  
    \end{tabular}
    \label{tab:similarity}
\end{table}

\subsection{Results on LongBench}
\begin{figure}[H]
    \centering
    \adjincludegraphics[width=0.99\linewidth, max width=0.99\columnwidth]{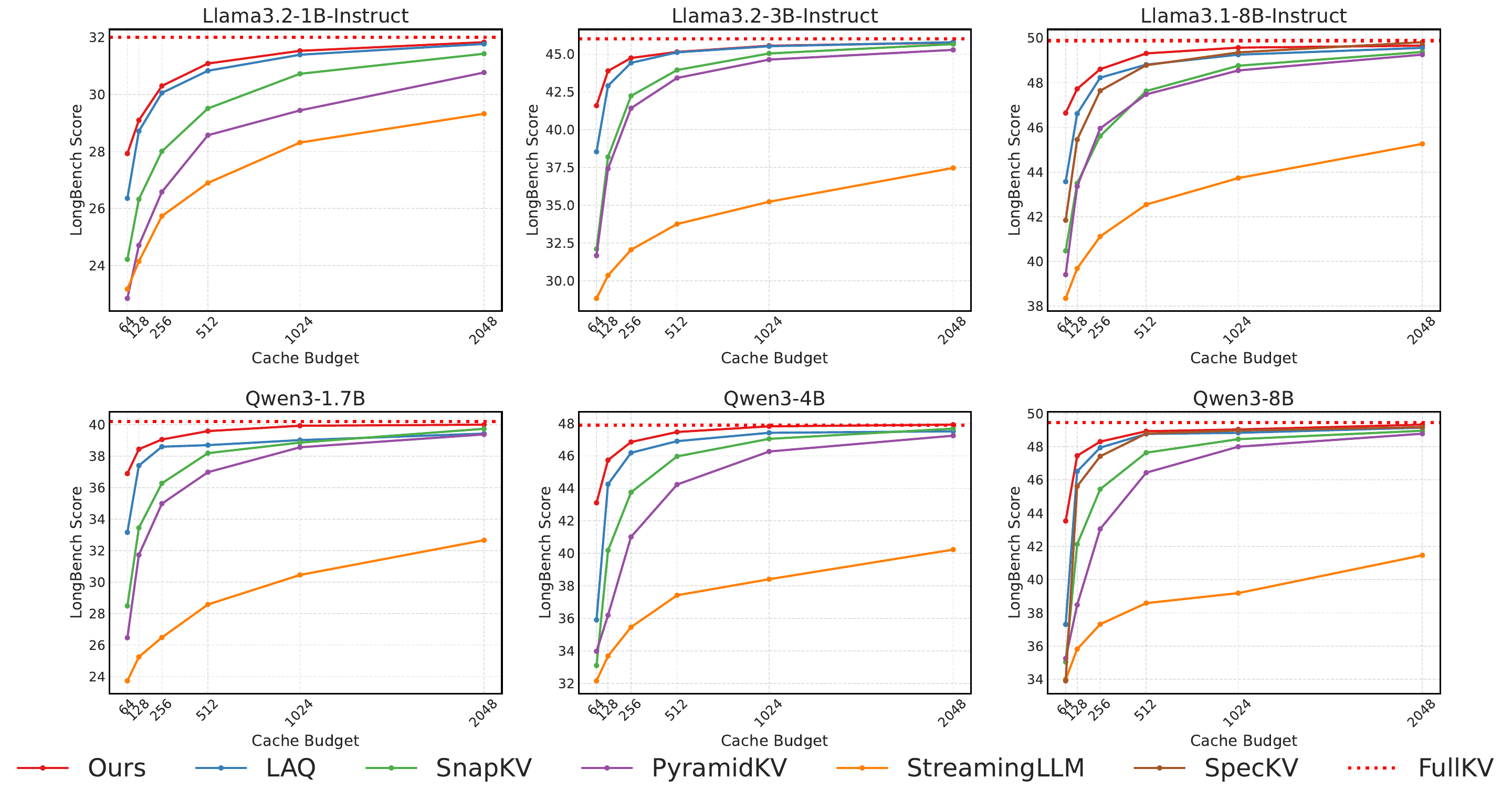}  
    \caption{
    Full Longbench results across multiple cache budgets. 1B-scale results are included.
    }\vspace{-0.3cm}
    \label{fig:longbench_main}
\end{figure}
\newpage
\begin{table*}[!ht]
    \centering
    \caption{LongBench evaluation results for Llama3.2-1B}
    \label{tab:longbench_llama1b}
    \resizebox{1.05\linewidth}{!}{%
    \fontsize{20}{20}\selectfont
    \begin{tabular}{llccccccccccccccccc}
        \toprule
        \multicolumn{2}{c}{\multirow{2}{*}{}} & \multicolumn{3}{c}{Single-Document QA} & \multicolumn{3}{c}{Multi-Document QA} & \multicolumn{3}{c}{Summarization} & \multicolumn{3}{c}{Few-shot Learning} & \multicolumn{2}{c}{Synthetic} & \multicolumn{2}{c}{Code} & \multirow{2}{*}{Avg.} \\
        \cmidrule(lr){3-5} \cmidrule(lr){6-8} \cmidrule(lr){9-11} \cmidrule(lr){12-14} \cmidrule(lr){15-16} \cmidrule(lr){17-18}

        & & NrtQA & Qasper & MF-en & HotpotQA & 2WikiMQA & Musique & GovReport & QMSum & MultiNews & TREC & TriviaQA & SAMSum & PCount & Pre & Lcc & RB-P & \\
        \midrule
        \midrule
  
        & FullKV                    & 19.24 & 15.96 & 42.47 & 35.53 & 29.42 & 19.87 & 28.34 & 22.18 & 25.64 & 64.00 & 80.85 & 38.83 & 3.00 & 4.78 & 38.63 & 43.35 & 32.01 \\

        & \multicolumn{18}{c}{\cellcolor{gray!25}\textit{KV Cache Size = 64}} \\
        & StreamingLLM              & 14.42 & 12.30 & 24.02 & 24.04 & 24.05 & 9.35  & 13.56 & 19.23 & 13.61 & 35.50 & 68.94 & 29.09 & 4.00 & 3.71 & 37.61 & 37.33 & 23.17 \\
        & SnapKV                    & 14.24 & 12.18 & 30.53 & 27.30 & 25.66 & 12.44 & 14.27 & 19.27 & 12.37 & 35.00 & 73.57 & 29.29 & 2.00 & 4.49 & 38.01 & 36.84 & 24.22 \\
        & PyramidKV                 & 13.33 & 11.36 & 26.05 & 25.34 & 24.01 & 10.57 & 13.64 & 19.49 & 11.96 & 35.00 & 69.95 & 27.75 & 1.50 & 4.33 & 35.25 & 36.01 & 22.85 \\
        & LAQ                       & 17.21 & 12.76 & 37.30 & 30.27 & 27.36 & 14.22 & 16.42 & 20.09 & 14.28 & 39.50 & 76.02 & 31.19 & 3.00 & 4.58 & 39.15 & 38.37 & 26.36 \\
        & \textbf{\lookaheadkv}      & 17.69 & 13.30 & 40.80 & 33.66 & 29.80 & 16.95 & 18.76 & 20.65 & 18.97 & 45.50 & 80.12 & 34.77 & 2.50 & 3.22 & 33.69 & 36.47 & \textbf{27.93} \\
        
        & \multicolumn{18}{c}{\cellcolor{gray!25}\textit{KV Cache Size = 128}} \\
        & StreamingLLM              & 14.84 & 12.36 & 24.67 & 25.48 & 23.86 & 8.73 & 14.71 & 19.58 & 15.50 & 38.00 & 71.61 & 31.82 & 3.50 & 3.79 & 38.81 & 39.03 & 24.14 \\  
        & SnapKV                    & 15.74 & 12.59 & 35.87 & 29.77 & 26.43 & 14.17 & 16.17 & 20.35 & 16.47 & 36.50 & 78.04 & 31.84 & 3.50 & 4.57 & 39.03 & 40.15 & 26.32 \\  
        & PyramidKV                 & 14.84 & 12.10 & 33.63 & 27.73 & 23.95 & 11.77 & 15.27 & 19.79 & 13.99 & 35.50 & 74.17 & 30.50 & 1.50 & 4.71 & 37.90 & 37.99 & 24.71 \\  
        & LAQ                       & 18.63 & 13.65 & 41.78 & 34.75 & 29.59 & 16.57 & 18.89 & 20.88 & 19.19 & 44.00 & 79.29 & 34.89 & 2.54 & 4.25 & 39.43 & 41.06 & 28.71 \\  
        & \textbf{\lookaheadkv}      & 17.38 & 14.92 & 41.39 & 35.46 & 29.22 & 17.47 & 20.13 & 20.78 & 21.24 & 51.50 & 80.27 & 36.19 & 3.00 & 4.17 & 34.75 & 37.68 & \textbf{29.10} \\ 

        & \multicolumn{18}{c}{\cellcolor{gray!25}\textit{KV Cache Size = 256}} \\
        & StreamingLLM              & 14.74 & 12.39 & 25.28 & 26.71 & 23.87 & 8.88 & 16.99 & 19.67 & 17.97 & 44.50 & 74.85 & 35.96 & 3.50 & 3.90 & 40.37 & 42.19 & 25.74 \\  
        & SnapKV                    & 16.59 & 13.78 & 38.80 & 32.54 & 28.11 & 16.55 & 18.55 & 20.00 & 19.69 & 41.50 & 79.31 & 33.70 & 4.00 & 4.58 & 39.15 & 41.26 & 28.01 \\  
        & PyramidKV                 & 15.11 & 13.08 & 37.31 & 32.03 & 25.36 & 12.60 & 16.91 & 20.10 & 17.78 & 40.50 & 76.61 & 32.34 & 3.50 & 4.65 & 37.13 & 40.32 & 26.58 \\  
        & LAQ                       & 18.31 & 14.64 & 41.83 & 35.34 & 29.61 & 17.27 & 20.68 & 21.21 & 21.37 & 52.00 & 79.62 & 36.99 & 4.04 & 4.17 & 40.44 & 43.37 & 30.06 \\  
        & \textbf{\lookaheadkv}      & 18.23 & 14.70 & 40.25 & 36.52 & 30.45 & 18.50 & 21.75 & 20.91 & 22.46 & 57.50 & 80.09 & 38.05 & 4.00 & 4.50 & 36.19 & 40.73 & \textbf{30.30} \\  

        & \multicolumn{18}{c}{\cellcolor{gray!25}\textit{KV Cache Size = 512}} \\
        & StreamingLLM              & 14.62 & 12.86 & 26.37 & 27.03 & 24.19 & 9.96  & 19.02 & 19.61 & 20.99 & 52.50 & 76.92 & 36.50 & 2.54 & 3.64 & 40.76 & 42.83 & 26.90 \\
        & SnapKV                    & 17.58 & 14.17 & 40.91 & 34.57 & 29.19 & 16.74 & 20.32 & 20.78 & 22.10 & 52.50 & 80.29 & 34.65 & 3.00 & 4.58 & 38.20 & 42.59 & 29.51 \\
        & PyramidKV                 & 16.55 & 13.48 & 39.67 & 32.62 & 28.38 & 15.59 & 18.50 & 20.87 & 20.54 & 48.50 & 79.30 & 34.43 & 4.00 & 4.50 & 38.79 & 41.42 & 28.57 \\
        & LAQ                       & 18.45 & 15.26 & 41.93 & 34.75 & 30.44 & 17.63 & 22.39 & 21.45 & 23.11 & 57.50 & 79.04 & 37.81 & 4.00 & 4.17 & 40.76 & 44.64 & 30.83 \\
        & \textbf{\lookaheadkv}      & 18.32 & 14.87 & 41.62 & 36.05 & 30.10 & 18.77 & 23.06 & 21.49 & 23.57 & 63.50 & 80.40 & 38.73 & 3.00 & 4.75 & 37.12 & 42.04 & \textbf{31.09} \\

        & \multicolumn{18}{c}{\cellcolor{gray!25}\textit{KV Cache Size = 1024}} \\
        & StreamingLLM            & 15.07 & 13.49 & 29.51 & 28.66 & 25.17 & 11.51 & 20.86 & 19.85 & 23.62 & 57.00 & 79.36 & 37.59 & 4.00 & 3.64 & 39.54 & 44.15 & 28.31 \\  
        & SnapKV                  & 18.12 & 14.66 & 40.52 & 35.01 & 29.85 & 18.86 & 22.38 & 21.12 & 24.16 & 60.50 & 81.16 & 35.61 & 3.00 & 4.62 & 38.58 & 43.47 & 30.73 \\  
        & PyramidKV               & 15.94 & 14.15 & 39.34 & 34.16 & 28.37 & 16.68 & 20.27 & 21.11 & 23.14 & 56.00 & 79.93 & 35.36 & 1.50 & 4.58 & 38.48 & 42.06 & 29.44 \\  
        & LAQ                     & 19.18 & 15.23 & 41.55 & 34.43 & 30.52 & 18.35 & 23.96 & 21.47 & 24.51 & 61.50 & 79.78 & 38.53 & 3.00 & 4.42 & 40.60 & 45.30 & 31.40 \\    
        & \textbf{\lookaheadkv}    & 18.51 & 15.41 & 41.49 & 35.41 & 29.74 & 19.29 & 24.93 & 21.19 & 24.58 & 64.00 & 81.07 & 39.06 & 3.50 & 4.80 & 37.99 & 43.55 & \textbf{31.53} \\

        & \multicolumn{18}{c}{\cellcolor{gray!25}\textit{KV Cache Size = 2048}} \\  
        & StreamingLLM          & 17.10 & 14.71 & 31.30 & 31.33 & 26.60 & 11.20 & 22.94 & 20.21 & 24.89 & 59.00 & 79.81 & 38.02 & 4.04 & 4.00 & 39.49 & 44.55 & 29.32 \\
        & SnapKV                & 17.73 & 15.74 & 42.03 & 36.12 & 29.48 & 19.34 & 24.30 & 21.75 & 25.22 & 62.50 & 80.90 & 38.22 & 3.00 & 4.75 & 38.28 & 43.52 & 31.43 \\
        & PyramidKV             & 18.83 & 14.50 & 41.40 & 35.75 & 28.89 & 17.40 & 22.05 & 21.14 & 24.97 & 60.50 & 80.62 & 37.33 & 2.50 & 4.50 & 38.51 & 43.46 & 30.77 \\
        & LAQ                   & 19.03 & 15.61 & 40.93 & 34.10 & 30.23 & 18.99 & 25.75 & 21.55 & 25.49 & 64.50 & 79.73 & 38.66 & 3.50 & 4.33 & 40.35 & 45.65 & 31.78 \\
        & \textbf{\lookaheadkv}  & 18.18 & 16.08 & 42.13 & 35.45 & 30.13 & 19.89 & 26.34 & 21.23 & 25.63 & 64.00 & 80.90 & 39.52 & 3.00 & 4.70 & 38.06 & 44.13 & \textbf{31.84} \\
        \cmidrule(lr){2-19}
        \bottomrule
    \end{tabular}
    } 
\end{table*}
\begin{table*}[!ht]
    \centering
    \caption{LongBench evaluation results for Qwen3-1.7B}
    \label{tab:longbench_qwen1b}
    \resizebox{1.05\linewidth}{!}{%
    \fontsize{20}{20}\selectfont
    \begin{tabular}{llccccccccccccccccc}
        \toprule
        \multicolumn{2}{c}{\multirow{2}{*}{}} & \multicolumn{3}{c}{Single-Document QA} & \multicolumn{3}{c}{Multi-Document QA} & \multicolumn{3}{c}{Summarization} & \multicolumn{3}{c}{Few-shot Learning} & \multicolumn{2}{c}{Synthetic} & \multicolumn{2}{c}{Code} & \multirow{2}{*}{Avg.} \\
        \cmidrule(lr){3-5} \cmidrule(lr){6-8} \cmidrule(lr){9-11} \cmidrule(lr){12-14} \cmidrule(lr){15-16} \cmidrule(lr){17-18}

        & & NrtQA & Qasper & MF-en & HotpotQA & 2WikiMQA & Musique & GovReport & QMSum & MultiNews & TREC & TriviaQA & SAMSum & PCount & Pre & Lcc & RB-P & \\
        \midrule
        \midrule
  
        & FullKV                    & 18.94 & 24.78 & 46.15 & 39.15 & 33.11 & 19.03 & 30.33 & 23.06 & 25.18 & 74.00 & 85.21 & 42.87 & 0.00 & 94.50 & 46.82 & 40.24 & 40.21 \\

        & \multicolumn{18}{c}{\cellcolor{gray!25}\textit{KV Cache Size = 64}} \\
        & StreamingLLM              & 12.25 & 16.83 & 20.38 & 22.62 & 24.61 & 7.26 & 11.43 & 18.96 & 12.12 & 39.00 & 66.48 & 35.14 & 0.00 & 12.50 & 42.95 & 36.90 & 23.71 \\  
        & SnapKV                    & 13.01 & 17.50 & 30.47 & 28.14 & 25.68 & 8.63 & 11.89 & 19.77 & 11.03 & 37.50 & 78.59 & 37.62 & 0.00 & 54.50 & 44.42 & 36.92 & 28.48 \\  
        & PyramidKV                 & 13.20 & 16.31 & 27.19 & 23.83 & 25.80 & 6.95 & 11.23 & 19.12 & 10.72 & 37.00 & 73.99 & 35.99 & 0.00 & 43.00 & 43.01 & 35.97 & 26.46 \\  
        & LAQ                       & 16.64 & 16.99 & 43.66 & 34.19 & 31.41 & 14.00 & 16.09 & 20.94 & 13.76 & 48.00 & 83.63 & 38.89 & 0.00 & 70.75 & 42.51 & 39.08 & 33.16 \\  
        & \textbf{\lookaheadkv}      & 19.12 & 21.73 & 44.25 & 37.18 & 33.01 & 15.98 & 21.42 & 22.53 & 19.73 & 56.50 & 85.56 & 41.72 & 0.00 & 89.00 & 44.58 & 38.05 & \textbf{36.90} \\ 
        
        & \multicolumn{18}{c}{\cellcolor{gray!25}\textit{KV Cache Size = 128}} \\
        & StreamingLLM              & 13.58 & 16.57 & 22.67 & 22.16 & 24.16 & 7.54 & 13.01 & 19.26 & 14.80 & 44.50 & 70.96 & 37.72 & 0.00 & 12.50 & 46.07 & 38.30 & 25.24 \\
        & SnapKV                    & 15.94 & 18.73 & 38.02 & 33.35 & 28.02 & 11.74 & 15.40 & 20.80 & 15.71 & 47.00 & 81.65 & 38.77 & 0.00 & 86.00 & 45.47 & 38.55 & 33.45 \\    
        & PyramidKV                 & 14.13 & 18.14 & 35.49 & 30.97 & 26.99 & 10.53 & 14.27 & 20.50 & 14.29 & 45.50 & 78.72 & 38.06 & 0.00 & 78.00 & 44.05 & 37.95 & 31.72 \\  
        & LAQ                       & 18.38 & 21.08 & 45.04 & 38.04 & 33.52 & 15.36 & 20.17 & 22.70 & 18.78 & 62.50 & 85.21 & 41.79 & 0.14 & 92.00 & 42.94 & 40.84 & 37.41 \\  
        & \textbf{\lookaheadkv}      & 19.46 & 23.27 & 44.59 & 37.81 & 33.82 & 17.97 & 23.71 & 23.12 & 21.70 & 65.50 & 85.56 & 42.38 & 0.12 & 92.75 & 44.80 & 38.61 & \textbf{38.45} \\  

        & \multicolumn{18}{c}{\cellcolor{gray!25}\textit{KV Cache Size = 256}} \\
        & StreamingLLM              & 13.41 & 17.66 & 22.58 & 23.72 & 23.88 & 7.65 & 16.15 & 19.24 & 17.97 & 47.50 & 76.22 & 40.17 & 0.00 & 13.50 & 46.27 & 37.64 & 26.47 \\  
        & SnapKV                    & 17.52 & 20.72 & 39.73 & 34.12 & 30.25 & 14.94 & 19.06 & 21.99 & 19.43 & 61.00 & 83.82 & 38.53 & 0.00 & 94.50 & 44.98 & 39.87 & 36.28 \\  
        & PyramidKV                 & 17.07 & 19.74 & 38.19 & 33.18 & 29.09 & 13.95 & 17.92 & 21.26 & 17.67 & 56.00 & 81.47 & 39.26 & 0.00 & 91.50 & 45.19 & 38.23 & 34.98 \\  
        & LAQ                       & 18.74 & 21.71 & 45.85 & 37.93 & 33.55 & 16.16 & 22.63 & 23.05 & 21.51 & 70.00 & 85.21 & 42.73 & 0.17 & 95.00 & 41.83 & 41.63 & 38.61 \\  
        & \textbf{\lookaheadkv}      & 19.60 & 24.30 & 45.69 & 38.81 & 34.02 & 17.91 & 25.51 & 23.11 & 23.15 & 70.00 & 85.37 & 42.16 & 0.12 & 91.00 & 45.29 & 38.98 & \textbf{39.06} \\  

        & \multicolumn{18}{c}{\cellcolor{gray!25}\textit{KV Cache Size = 512}} \\
        & StreamingLLM          & 13.92 & 18.40 & 25.10 & 24.91 & 24.77 & 7.67 & 20.38 & 19.63 & 20.78 & 61.00 & 81.80 & 40.38 & 0.00 & 12.50 & 47.28 & 38.62 & 28.57 \\ 
        & SnapKV                & 19.32 & 22.22 & 44.07 & 36.25 & 30.04 & 15.66 & 22.20 & 22.15 & 21.73 & 70.50 & 84.81 & 40.54 & 0.14 & 94.50 & 46.75 & 40.26 & 38.20 \\  
        & PyramidKV             & 17.95 & 20.69 & 41.43 & 36.22 & 29.68 & 14.96 & 20.39 & 21.52 & 20.01 & 66.00 & 84.65 & 40.16 & 0.14 & 93.50 & 45.61 & 38.83 & 36.98 \\  
        & LAQ                   & 16.99 & 22.67 & 46.97 & 38.10 & 33.62 & 16.38 & 24.58 & 23.39 & 22.99 & 71.50 & 84.71 & 42.25 & 0.00 & 94.00 & 40.85 & 40.30 & 38.71 \\  
        & \textbf{\lookaheadkv}  & 19.04 & 24.66 & 44.68 & 39.04 & 33.66 & 17.64 & 27.46 & 23.31 & 24.17 & 73.00 & 85.37 & 42.87 & 0.17 & 94.00 & 45.27 & 39.25 & \textbf{39.60} \\  

        & \multicolumn{18}{c}{\cellcolor{gray!25}\textit{KV Cache Size = 1024}} \\
        & StreamingLLM          & 15.32 & 18.63 & 26.90 & 27.88 & 26.44 & 8.47 & 23.56 & 20.34 & 23.82 & 65.50 & 84.00 & 41.69 & 0.00 & 18.00 & 46.93 & 39.71 & 30.45 \\  
        & SnapKV                & 18.68 & 24.05 & 44.25 & 38.57 & 30.72 & 16.63 & 24.97 & 22.28 & 23.62 & 71.50 & 85.26 & 40.43 & 0.50 & 95.00 & 46.39 & 39.03 & 38.87 \\  
        & PyramidKV             & 18.76 & 23.44 & 43.47 & 36.96 & 29.98 & 16.21 & 23.08 & 22.21 & 22.68 & 72.50 & 84.76 & 40.19 & 0.50 & 95.50 & 46.89 & 39.98 & 38.57 \\  
        & LAQ                   & 16.76 & 22.60 & 45.45 & 38.65 & 33.44 & 17.15 & 26.68 & 23.32 & 24.11 & 73.00 & 84.71 & 43.11 & 0.00 & 95.00 & 39.42 & 41.02 & 39.03 \\  
        & \textbf{\lookaheadkv}  & 19.20 & 25.15 & 44.48 & 39.09 & 32.85 & 18.39 & 29.18 & 23.24 & 25.14 & 73.00 & 84.84 & 43.19 & 0.17 & 95.00 & 46.24 & 39.86 & \textbf{39.94} \\  

        & \multicolumn{18}{c}{\cellcolor{gray!25}\textit{KV Cache Size = 2048}} \\  
        & StreamingLLM          & 16.32 & 22.12 & 29.96 & 31.34 & 28.13 & 9.37 & 26.32 & 20.96 & 24.58 & 68.50 & 84.59 & 42.84 & 0.00 & 31.00 & 46.63 & 39.92 & 32.66 \\  
        & SnapKV                & 19.44 & 24.89 & 45.18 & 38.97 & 32.65 & 17.32 & 27.89 & 22.60 & 24.81 & 72.50 & 85.21 & 42.37 & 0.17 & 95.00 & 46.50 & 40.43 & 39.75 \\  
        & PyramidKV             & 19.52 & 24.01 & 44.53 & 39.48 & 31.94 & 16.37 & 26.16 & 22.68 & 24.68 & 72.00 & 84.84 & 41.79 & 0.50 & 95.00 & 46.57 & 40.12 & 39.39 \\  
        & LAQ                   & 16.42 & 22.65 & 45.81 & 38.66 & 33.44 & 17.38 & 28.79 & 23.41 & 25.14 & 73.00 & 84.71 & 42.94 & 0.00 & 95.00 & 43.38 & 40.30 & 39.44 \\  
        & \textbf{\lookaheadkv}  & 19.08 & 25.15 & 45.04 & 38.79 & 33.00 & 17.85 & 29.86 & 23.00 & 25.26 & 73.50 & 85.21 & 43.06 & 0.17 & 94.00 & 46.62 & 40.53 & \textbf{40.01} \\      
        \cmidrule(lr){2-19}
        \bottomrule
    \end{tabular}
    } 
\end{table*}   
\newpage
\begin{table*}[!ht]
    \centering
    \caption{LongBench evaluation results for Llama3.2-3B}
    \label{tab:longbench_llama3b}
    \resizebox{1.05\linewidth}{!}{%
    \fontsize{20}{20}\selectfont
    \begin{tabular}{llccccccccccccccccc}
        \toprule
        \multicolumn{2}{c}{\multirow{2}{*}{}} & \multicolumn{3}{c}{Single-Document QA} & \multicolumn{3}{c}{Multi-Document QA} & \multicolumn{3}{c}{Summarization} & \multicolumn{3}{c}{Few-shot Learning} & \multicolumn{2}{c}{Synthetic} & \multicolumn{2}{c}{Code} & \multirow{2}{*}{Avg.} \\
        \cmidrule(lr){3-5} \cmidrule(lr){6-8} \cmidrule(lr){9-11} \cmidrule(lr){12-14} \cmidrule(lr){15-16} \cmidrule(lr){17-18}

        & & NrtQA & Qasper & MF-en & HotpotQA & 2WikiMQA & Musique & GovReport & QMSum & MultiNews & TREC & TriviaQA & SAMSum & PCount & Pre & Lcc & RB-P & \\
        \midrule
        \midrule
  
        & FullKV                    & 25.35 & 40.77 & 50.31 & 53.79 & 40.05 & 25.62 & 33.11 & 24.38 & 25.95 & 72.00 & 88.89 & 43.25 & 4.00 & 97.50 & 54.65 & 56.83 & 46.03  \\

        & \multicolumn{18}{c}{\cellcolor{gray!25}\textit{KV Cache Size = 64}} \\
        & StreamingLLM              & 19.79 & 20.68 & 24.68 & 42.63 & 34.09 & 16.14 & 16.53 & 20.22 & 14.78 & 38.00 & 75.84 & 33.19 & 4.50 & 10.50 & 46.11 & 43.76 & 28.84 \\
        & SnapKV                    & 21.88 & 20.56 & 39.12 & 48.33 & 35.39 & 22.39 & 16.37 & 20.30 & 14.41 & 39.50 & 82.17 & 35.05 & 4.50 & 21.50 & 47.96 & 44.30 & 32.11 \\
        & PyramidKV                 & 23.06 & 19.20 & 36.26 & 46.34 & 34.81 & 20.17 & 16.17 & 20.96 & 14.02 & 41.00 & 81.93 & 35.08 & 4.00 & 23.00 & 46.82 & 43.94 & 31.67  \\
        & LAQ                       & 23.33 & 28.03 & 48.16 & 54.08 & 37.28 & 23.91 & 18.88 & 21.75 & 16.62 & 47.50 & 86.65 & 37.73 & 4.50 & 74.00 & 47.82 & 46.56 & 38.55  \\
        & \textbf{\lookaheadkv}      & 25.22 & 33.43 & 49.90 & 53.50 & 39.60 & 24.31 & 22.42 & 22.23 & 20.30 & 64.00 & 89.38 & 38.70 & 4.50 & 77.50 & 50.79 & 49.83 & \textbf{41.60} \\
        
        & \multicolumn{18}{c}{\cellcolor{gray!25}\textit{KV Cache Size = 128}} \\
        & StreamingLLM              & 19.27 & 20.13 & 25.60 & 43.77 & 33.92 & 16.84 & 17.52 & 19.96 & 17.23 & 42.50 & 79.63 & 36.86 & 4.00 & 10.50 & 50.75 & 47.3 & 30.36  \\ 
        & SnapKV                    & 20.59 & 27.04 & 45.17 & 51.15 & 37.77 & 21.49 & 19.64 & 21.89 & 18.22 & 49.50 & 86.62 & 37.35 & 5.50 & 68.00 & 51.51 & 49.81 & 38.20 \\  
        & PyramidKV                 & 22.19 & 25.35 & 45.65 & 51.67 & 37.41 & 20.93 & 19.29 & 21.45 & 18.37 & 47.50 & 84.95 & 37.59 & 5.50 & 62.00 & 50.54 & 48.46 & 37.43 \\  
        & LAQ                       & 25.04 & 35.33 & 51.55 & 53.87 & 39.98 & 23.75 & 21.82 & 23.27 & 20.20 & 61.00 & 89.13 & 39.91 & 5.53 & 89.50 & 53.67 & 53.08 & 42.91 \\  
        & \textbf{\lookaheadkv}      & 25.45 & 36.54 & 51.58 & 53.30 & 40.47 & 24.38 & 24.02 & 23.64 & 21.95 & 68.50 & 89.56 & 41.64 & 4.00 & 91.00 & 53.29 & 53.06 & \textbf{43.90} \\ 

        & \multicolumn{18}{c}{\cellcolor{gray!25}\textit{KV Cache Size = 256}} \\
        & StreamingLLM              & 19.42 & 22.77 & 28.79 & 43.12 & 32.29 & 16.69 & 19.61 & 19.78 & 19.15 & 50.00 & 84.07 & 39.15 & 4.00 & 11.50 & 52.76 & 49.85 & 32.06 \\   
        & SnapKV                    & 22.44 & 30.24 & 47.71 & 53.97 & 40.23 & 23.01 & 21.73 & 22.52 & 21.03 & 61.50 & 88.81 & 38.69 & 5.50 & 91.50 & 53.49 & 53.56 & 42.25 \\ 
        & PyramidKV                 & 21.35 & 30.12 & 47.93 & 53.26 & 38.78 & 23.27 & 21.26 & 22.61 & 20.57 & 61.00 & 87.58 & 39.01 & 5.50 & 87.00 & 52.95 & 50.72 & 41.43 \\ 
        & LAQ                       & 24.46 & 38.11 & 50.89 & 53.80 & 39.58 & 24.09 & 23.61 & 23.83 & 22.10 & 70.00 & 89.36 & 41.30 & 4.03 & 95.00 & 55.55 & 55.21 & 44.43 \\  
        & \textbf{\lookaheadkv}      & 25.58 & 37.51 & 50.92 & 53.20 & 40.35 & 25.74 & 25.92 & 24.37 & 23.36 & 69.00 & 89.51 & 42.54 & 4.50 & 94.50 & 53.92 & 55.20 & \textbf{44.76} \\  

        & \multicolumn{18}{c}{\cellcolor{gray!25}\textit{KV Cache Size = 512}} \\
        & StreamingLLM              & 19.42 & 23.98 & 29.39 & 43.87 & 32.82 & 16.92 & 21.95 & 20.37 & 22.14 & 59.50 & 85.32 & 41.04 & 4.00 & 12.50 & 55.06 & 51.93 & 33.76 \\
        & SnapKV                    & 23.50 & 35.97 & 49.42 & 52.38 & 39.49 & 23.50 & 23.68 & 23.47 & 22.97 & 68.50 & 89.16 & 40.43 & 4.00 & 97.00 & 55.01 & 54.87 & 43.96 \\  
        & PyramidKV                 & 22.85 & 34.18 & 48.20 & 52.36 & 39.98 & 23.36 & 22.98 & 23.38 & 22.48 & 68.50 & 88.41 & 39.90 & 4.50 & 96.50 & 53.68 & 53.67 & 43.43 \\  
        & LAQ                       & 24.90 & 39.15 & 50.66 & 53.88 & 39.85 & 26.16 & 25.45 & 23.93 & 23.88 & 71.00 & 88.97 & 42.57 & 4.00 & 96.50 & 55.48 & 55.68 & 45.13 \\  
        & \textbf{\lookaheadkv}      & 24.12 & 38.81 & 50.78 & 53.84 & 40.06 & 25.28 & 27.81 & 24.15 & 24.83 & 69.00 & 89.16 & 42.01 & 4.00 & 97.00 & 55.03 & 56.64 & \textbf{45.16} \\  

        & \multicolumn{18}{c}{\cellcolor{gray!25}\textit{KV Cache Size = 1024}} \\
        & StreamingLLM            & 21.33 & 27.49 & 31.93 & 44.76 & 35.00 & 16.63 & 24.32 & 21.01 & 24.45 & 61.50 & 85.48 & 40.93 & 4.00 & 16.50 & 55.09 & 53.40 & 35.24 \\  
        & SnapKV                  & 24.67 & 38.16 & 51.18 & 54.02 & 40.13 & 25.01 & 25.92 & 23.78 & 24.30 & 69.50 & 88.79 & 40.79 & 4.50 & 98.00 & 55.57 & 56.58 & 45.06 \\  
        & PyramidKV               & 24.01 & 37.98 & 51.43 & 53.64 & 39.49 & 24.15 & 25.48 & 23.49 & 24.50 & 68.50 & 88.68 & 40.76 & 4.00 & 98.00 & 55.14 & 55.15 & 44.65 \\  
        & LAQ                     & 24.55 & 40.78 & 49.15 & 53.65 & 40.01 & 26.68 & 27.69 & 24.07 & 25.25 & 71.50 & 88.89 & 43.54 & 4.00 & 97.00 & 54.93 & 56.88 & 45.54 \\  
        & \textbf{\lookaheadkv}    & 24.28 & 39.51 & 51.35 & 53.64 & 40.36 & 25.07 & 29.61 & 24.17 & 25.41 & 69.50 & 88.99 & 42.90 & 4.00 & 98.00 & 55.59 & 56.72 & \textbf{45.57} \\  

        & \multicolumn{18}{c}{\cellcolor{gray!25}\textit{KV Cache Size = 2048}} \\  
        & StreamingLLM          & 22.36 & 32.37 & 33.12 & 46.79 & 36.00 & 17.95 & 26.86 & 21.68 & 25.78 & 66.00 & 87.26 & 41.68 & 4.00 & 27.50 & 55.48 & 54.82 & 37.48 \\
        & SnapKV                & 24.54 & 40.50 & 51.36 & 53.80 & 40.60 & 25.24 & 28.41 & 23.92 & 25.62 & 71.50 & 88.86 & 42.39 & 4.00 & 98.00 & 55.07 & 56.97 & 45.67 \\    
        & PyramidKV             & 24.05 & 39.80 & 51.31 & 53.88 & 40.03 & 24.50 & 27.88 & 24.27 & 25.65 & 70.00 & 88.80 & 42.11 & 4.00 & 98.00 & 54.84 & 55.59 & 45.29 \\  
        & LAQ                   & 24.80 & 41.44 & 49.50 & 54.05 & 39.88 & 26.13 & 30.09 & 24.44 & 25.80 & 71.50 & 88.89 & 43.11 & 4.00 & 97.00 & 55.44 & 56.77 & \textbf{45.80} \\  
        & \textbf{\lookaheadkv}  & 25.23 & 40.67 & 50.00 & 53.79 & 40.06 & 24.87 & 31.30 & 24.18 & 25.88 & 71.00 & 88.99 & 43.49 & 4.00 & 97.00 & 55.12 & 56.77 & 45.77 \\
        \cmidrule(lr){2-19}
        \bottomrule
    \end{tabular}
    } 
\end{table*}
\begin{table*}[!ht]
    \centering
    \caption{LongBench evaluation results for Qwen3-4B}
    \label{tab:longbench_qwen4b}
    \resizebox{1.05\linewidth}{!}{%
    \fontsize{20}{20}\selectfont
    \begin{tabular}{llccccccccccccccccc}
        \toprule
        \multicolumn{2}{c}{\multirow{2}{*}{}} & \multicolumn{3}{c}{Single-Document QA} & \multicolumn{3}{c}{Multi-Document QA} & \multicolumn{3}{c}{Summarization} & \multicolumn{3}{c}{Few-shot Learning} & \multicolumn{2}{c}{Synthetic} & \multicolumn{2}{c}{Code} & \multirow{2}{*}{Avg.} \\
        \cmidrule(lr){3-5} \cmidrule(lr){6-8} \cmidrule(lr){9-11} \cmidrule(lr){12-14} \cmidrule(lr){15-16} \cmidrule(lr){17-18}

        & & NrtQA & Qasper & MF-en & HotpotQA & 2WikiMQA & Musique & GovReport & QMSum & MultiNews & TREC & TriviaQA & SAMSum & PCount & Pre & Lcc & RB-P & \\
        \midrule
        \midrule
  
        & FullKV & 27.45 & 43.30 & 54.45 & 55.63 & 43.43 & 31.61 & 32.24 & 24.61 & 25.00 & 73.00 & 88.76 & 43.65 & 0.75 & 96.50 & 64.29 & 61.39 & 47.88 \\

        & \multicolumn{18}{c}{\cellcolor{gray!25}\textit{KV Cache Size = 64}} \\
        & StreamingLLM          & 12.46 & 23.96 & 25.93 & 38.56 & 33.40 & 19.47 & 13.74 & 19.71 & 13.04 & 39.50 & 75.48 & 34.33 & 0.50 & 64.50 & 51.42 & 48.46 & 32.15 \\  
        & SnapKV                & 15.28 & 25.03 & 31.61 & 40.00 & 34.95 & 18.83 & 12.88 & 19.78 & 12.49 & 40.50 & 75.62 & 33.69 & 1.00 & 69.00 & 51.48 & 47.38 & 33.10 \\  
        & PyramidKV             & 15.50 & 24.84 & 34.33 & 40.70 & 35.07 & 19.39 & 13.48 & 19.85 & 13.03 & 41.50 & 76.69 & 33.95 & 1.50 & 73.00 & 52.98 & 47.77 & 33.97 \\  
        & LAQ                   & 16.55 & 30.74 & 46.21 & 40.58 & 38.10 & 18.35 & 14.96 & 20.74 & 14.48 & 43.50 & 71.25 & 34.40 & 1.50 & 81.25 & 53.45 & 48.45 & 35.91 \\  
        & \textbf{\lookaheadkv}  & 20.49 & 37.99 & 51.37 & 54.71 & 42.30 & 30.90 & 22.10 & 22.98 & 18.71 & 58.50 & 88.85 & 39.71 & 1.00 & 92.00 & 55.89 & 52.33 & \textbf{43.11} \\   
        
        & \multicolumn{18}{c}{\cellcolor{gray!25}\textit{KV Cache Size = 128}} \\
        & StreamingLLM          & 15.69 & 23.56 & 26.02 & 38.03 & 32.38 & 18.86 & 14.79 & 19.67 & 15.20 & 45.50 & 78.32 & 37.32 & 0.50 & 65.50 & 55.83 & 51.76 & 33.68 \\
        & SnapKV                & 19.56 & 29.48 & 43.55 & 49.81 & 37.95 & 25.99 & 16.27 & 21.17 & 16.16 & 49.50 & 86.31 & 38.10 & 1.50 & 95.00 & 58.75 & 53.78 & 40.18 \\
        & PyramidKV             & 15.88 & 27.32 & 38.51 & 40.25 & 33.56 & 20.45 & 14.72 & 20.88 & 14.65 & 45.00 & 76.75 & 35.06 & 1.50 & 89.50 & 55.58 & 49.49 & 36.19 \\
        & LAQ                   & 21.44 & 37.82 & 53.26 & 54.98 & 42.75 & 32.08 & 20.44 & 23.69 & 18.86 & 60.50 & 87.55 & 40.48 & 2.50 & 93.00 & 60.85 & 57.87 & 44.25 \\
        & \textbf{\lookaheadkv}  & 25.17 & 40.13 & 52.28 & 55.10 & 43.47 & 31.38 & 24.83 & 24.46 & 21.57 & 67.00 & 88.85 & 41.37 & 1.00 & 96.50 & 61.00 & 57.77 & \textbf{45.74} \\  

        & \multicolumn{18}{c}{\cellcolor{gray!25}\textit{KV Cache Size = 256}} \\
        & StreamingLLM          & 15.66 & 25.74 & 28.99 & 37.34 & 32.47 & 19.02 & 17.65 & 20.12 & 18.02 & 49.50 & 81.97 & 38.87 & 1.00 & 66.50 & 59.45 & 55.05 & 35.46 \\
        & SnapKV                & 24.64 & 35.80 & 47.67 & 54.45 & 40.78 & 29.60 & 19.84 & 22.75 & 19.68 & 60.00 & 87.64 & 39.46 & 1.00 & 96.00 & 62.57 & 58.24 & 43.76 \\
        & PyramidKV             & 18.31 & 31.30 & 44.14 & 51.08 & 36.87 & 25.14 & 18.68 & 22.04 & 17.78 & 56.50 & 85.53 & 38.77 & 1.50 & 95.50 & 59.16 & 53.92 & 41.01 \\
        & LAQ                   & 26.88 & 40.94 & 53.82 & 55.76 & 43.22 & 31.53 & 23.34 & 24.03 & 21.57 & 68.50 & 87.72 & 41.61 & 2.00 & 93.50 & 62.60 & 62.03 & 46.19 \\
        & \textbf{\lookaheadkv}  & 26.25 & 41.08 & 53.03 & 55.21 & 43.28 & 31.99 & 27.13 & 25.09 & 23.46 & 71.50 & 88.76 & 41.89 & 1.00 & 96.50 & 63.42 & 60.09 & \textbf{46.85} \\  

        & \multicolumn{18}{c}{\cellcolor{gray!25}\textit{KV Cache Size = 512}} \\
        & StreamingLLM          & 18.02 & 27.64 & 30.11 & 39.03 & 33.32 & 20.70 & 21.47 & 20.39 & 21.96 & 60.50 & 85.45 & 40.27 & 0.50 & 59.50 & 62.54 & 57.33 & 37.42 \\  
        & SnapKV                & 25.27 & 39.10 & 51.45 & 54.22 & 42.21 & 32.86 & 23.30 & 23.53 & 22.33 & 70.00 & 88.76 & 40.24 & 1.00 & 96.50 & 64.28 & 60.45 & 45.97 \\  
        & PyramidKV             & 21.93 & 34.53 & 49.40 & 53.99 & 40.38 & 30.21 & 21.87 & 22.72 & 20.77 & 67.00 & 88.24 & 40.05 & 1.00 & 96.50 & 61.47 & 57.70 & 44.24 \\  
        & LAQ                   & 26.50 & 42.56 & 53.88 & 55.24 & 43.25 & 32.14 & 25.92 & 24.46 & 23.42 & 73.00 & 87.72 & 42.94 & 1.50 & 93.50 & 62.99 & 61.47 & 46.91 \\  
        & \textbf{\lookaheadkv}  & 26.86 & 41.97 & 53.10 & 55.59 & 43.97 & 32.09 & 29.57 & 25.35 & 24.61 & 72.00 & 88.76 & 42.85 & 1.50 & 96.50 & 63.83 & 60.96 & \textbf{47.47} \\ 

        & \multicolumn{18}{c}{\cellcolor{gray!25}\textit{KV Cache Size = 1024}} \\
        & StreamingLLM          & 20.48 & 30.08 & 32.30 & 42.20 & 34.23 & 20.65 & 24.81 & 20.84 & 24.19 & 64.50 & 87.39 & 40.95 & 1.00 & 47.00 & 64.74 & 59.17 & 38.41 \\
        & SnapKV                & 25.91 & 41.72 & 52.26 & 56.50 & 43.15 & 32.08 & 26.69 & 24.53 & 24.02 & 71.50 & 88.76 & 41.77 & 1.00 & 96.50 & 64.46 & 61.91 & 47.05 \\
        & PyramidKV             & 25.81 & 39.40 & 51.89 & 53.26 & 42.26 & 32.08 & 25.11 & 23.72 & 23.61 & 70.00 & 88.76 & 41.10 & 1.00 & 96.50 & 63.93 & 61.88 & 46.27 \\
        & LAQ                   & 27.40 & 43.93 & 54.30 & 55.95 & 43.62 & 31.66 & 28.18 & 25.16 & 24.64 & 73.00 & 87.77 & 43.33 & 1.75 & 93.50 & 62.54 & 62.00 & 47.42 \\
        & \textbf{\lookaheadkv}  & 27.47 & 42.45 & 53.70 & 55.64 & 43.85 & 32.40 & 30.68 & 24.98 & 25.21 & 73.00 & 88.76 & 42.96 & 1.00 & 96.50 & 64.61 & 61.89 & \textbf{47.82} \\ 

        & \multicolumn{18}{c}{\cellcolor{gray!25}\textit{KV Cache Size = 2048}} \\  
        & StreamingLLM          & 20.87 & 34.01 & 36.39 & 44.11 & 37.06 & 21.93 & 28.06 & 21.64 & 25.16 & 67.50 & 88.39 & 41.55 & 0.50 & 52.00 & 63.58 & 60.98 & 40.23 \\  
        & SnapKV                & 26.80 & 43.04 & 53.50 & 55.54 & 44.01 & 33.33 & 29.49 & 24.64 & 24.86 & 73.00 & 88.76 & 41.94 & 1.25 & 96.50 & 64.10 & 62.08 & 47.68 \\  
        & PyramidKV             & 25.74 & 42.42 & 53.91 & 55.34 & 43.12 & 33.06 & 27.70 & 24.21 & 24.74 & 72.00 & 88.76 & 41.54 & 1.25 & 96.50 & 63.81 & 61.78 & 47.24 \\  
        & LAQ                   & 27.21 & 43.52 & 53.62 & 55.67 & 43.89 & 31.73 & 30.42 & 24.93 & 25.04 & 73.00 & 87.72 & 43.77 & 1.50 & 93.25 & 63.02 & 61.92 & 47.51 \\  
        & \textbf{\lookaheadkv}  & 27.48 & 42.86 & 53.71 & 55.31 & 43.82 & 32.42 & 31.79 & 24.75 & 25.33 & 73.00 & 88.76 & 43.34 & 1.25 & 96.50 & 64.18 & 62.23 & \textbf{47.92} \\      
        \cmidrule(lr){2-19}
        \bottomrule
    \end{tabular}
    } 
\end{table*}
\newpage
\begin{table*}[!ht]
    \centering
    \caption{LongBench evaluation results for Llama3.1-8B}
    \label{tab:longbench_llama8b}
    \resizebox{1.05\linewidth}{!}{%
    \fontsize{20}{20}\selectfont
    \begin{tabular}{llccccccccccccccccc}
        \toprule
        \multicolumn{2}{c}{\multirow{2}{*}{}} & \multicolumn{3}{c}{Single-Document QA} & \multicolumn{3}{c}{Multi-Document QA} & \multicolumn{3}{c}{Summarization} & \multicolumn{3}{c}{Few-shot Learning} & \multicolumn{2}{c}{Synthetic} & \multicolumn{2}{c}{Code} & \multirow{2}{*}{Avg.} \\
        \cmidrule(lr){3-5} \cmidrule(lr){6-8} \cmidrule(lr){9-11} \cmidrule(lr){12-14} \cmidrule(lr){15-16} \cmidrule(lr){17-18}

        & & NrtQA & Qasper & MF-en & HotpotQA & 2WikiMQA & Musique & GovReport & QMSum & MultiNews & TREC & TriviaQA & SAMSum & PCount & Pre & Lcc & RB-P & \\
        \midrule
        \midrule
  
        & FullKV                & 31.63 & 46.66 & 56.93 & 58.10 & 48.50 & 31.57 & 34.46 & 25.28 & 26.98 & 72.50 & 91.65 & 43.79 & 6.64 & 99.50 & 65.12 & 58.78 & 49.88 \\

        & \multicolumn{18}{c}{\cellcolor{gray!25}\textit{KV Cache Size = 64}} \\
        & StreamingLLM          & 25.75 & 21.75 & 31.22 & 49.09 & 42.11 & 23.98 & 17.29 & 20.99 & 16.04 & 38.50 & 82.81 & 34.50 & 7.50 & 99.50 & 54.27 & 48.14 & 38.34 \\  
        & SnapKV                & 27.37 & 24.99 & 41.77 & 54.27 & 45.27 & 27.52 & 16.75 & 21.73 & 16.32 & 39.00 & 86.32 & 36.58 & 7.50 & 98.50 & 55.57 & 48.07 & 40.47 \\  
        & PyramidKV             & 24.25 & 22.87 & 41.03 & 53.07 & 43.55 & 26.36 & 16.46 & 21.52 & 15.61 & 38.50 & 81.95 & 36.68 & 7.50 & 99.50 & 54.40 & 47.20 & 39.40 \\  
        & LAQ                   & 27.62 & 33.71 & 52.35 & 55.85 & 48.92 & 28.06 & 19.74 & 23.19 & 18.90 & 46.00 & 88.29 & 40.62 & 6.83 & 100.00 & 55.55 & 51.49 & 43.57 \\  
        & SpecKV                & 24.87 & 26.57 & 51.22 & 55.29 & 46.57 & 25.42 & 19.78 & 22.29 & 19.20 & 33.50 & 85.12 & 39.14 & 8.50 & 97.00 & 57.78 & 57.19 & 41.84 \\  
        & \textbf{\lookaheadkv}  & 30.62 & 41.46 & 55.77 & 56.42 & 48.56 & 30.30 & 23.54 & 24.08 & 21.23 & 60.50 & 91.62 & 42.56 & 7.50 & 99.50 & 58.74 & 53.86 & \textbf{46.64} \\ 
        
        & \multicolumn{18}{c}{\cellcolor{gray!25}\textit{KV Cache Size = 128}} \\
        & StreamingLLM          & 24.95 & 21.50 & 32.56 & 50.67 & 42.89 & 24.31 & 18.49 & 21.25 & 18.19 & 40.50 & 85.57 & 38.28 & 7.50 & 99.50 & 59.03 & 49.72 & 39.68 \\  
        & SnapKV                & 29.13 & 28.06 & 51.23 & 56.79 & 45.30 & 27.81 & 19.99 & 23.03 & 19.73 & 46.00 & 89.72 & 40.44 & 7.50 & 99.50 & 59.50 & 52.19 & 43.50 \\  
        & PyramidKV             & 27.70 & 28.86 & 52.00 & 56.76 & 46.11 & 28.13 & 19.86 & 22.81 & 20.03 & 44.50 & 88.41 & 39.73 & 7.50 & 99.50 & 59.84 & 51.96 & 43.36 \\  
        & LAQ                   & 30.48 & 38.31 & 55.73 & 57.50 & 49.13 & 29.67 & 22.42 & 24.20 & 21.59 & 60.50 & 92.09 & 41.04 & 7.25 & 99.50 & 60.54 & 55.83 & 46.61 \\  
        & SpecKV                & 29.22 & 29.12 & 54.05 & 56.54 & 46.30 & 29.90 & 22.65 & 23.18 & 21.25 & 52.00 & 90.02 & 42.14 & 8.83 & 99.50 & 61.11 & 61.38 & 45.45 \\  
        & \textbf{\lookaheadkv}  & 31.32 & 42.85 & 56.78 & 57.04 & 47.44 & 30.82 & 25.18 & 24.33 & 23.09 & 65.50 & 92.24 & 42.96 & 7.50 & 99.50 & 61.75 & 55.29 & \textbf{47.72} \\ 

        & \multicolumn{18}{c}{\cellcolor{gray!25}\textit{KV Cache Size = 256}} \\
        & StreamingLLM          & 25.96 & 24.08 & 33.73 & 50.56 & 42.61 & 23.49 & 20.86 & 21.60 & 20.64 & 46.00 & 87.50 & 41.09 & 7.50 & 99.50 & 61.19 & 51.53 & 41.12 \\  
        & SnapKV                & 27.96 & 34.49 & 55.07 & 57.40 & 46.57 & 29.50 & 22.49 & 23.51 & 22.42 & 54.00 & 91.10 & 40.61 & 7.33 & 99.50 & 62.48 & 55.36 & 45.61 \\  
        & PyramidKV             & 28.09 & 36.64 & 55.86 & 57.68 & 46.28 & 29.56 & 22.23 & 23.86 & 22.53 & 56.50 & 91.56 & 41.23 & 7.33 & 99.50 & 62.47 & 53.92 & 45.95 \\  
        & LAQ                   & 31.03 & 43.97 & 55.93 & 57.78 & 49.42 & 30.42 & 24.48 & 24.60 & 23.29 & 68.00 & 92.20 & 42.61 & 7.08 & 100.00 & 62.70 & 58.09 & 48.23 \\  
        & SpecKV                & 28.66 & 36.19 & 57.26 & 58.17 & 48.51 & 30.85 & 24.83 & 24.60 & 23.32 & 61.00 & 91.16 & 42.46 & 8.33 & 99.50 & 64.21 & 63.18 & 47.64 \\  
        & \textbf{\lookaheadkv}  & 31.96 & 44.01 & 56.80 & 57.99 & 47.41 & 31.46 & 27.26 & 24.56 & 24.59 & 69.00 & 92.55 & 42.93 & 7.33 & 100.00 & 62.81 & 57.02 & \textbf{48.61} \\  

        & \multicolumn{18}{c}{\cellcolor{gray!25}\textit{KV Cache Size = 512}} \\
        & StreamingLLM          & 27.20 & 26.66 & 34.51 & 50.04 & 42.70 & 23.35 & 23.33 & 21.35 & 23.51 & 57.50 & 87.68 & 41.87 & 7.50 & 97.50 & 62.34 & 53.63 & 42.54 \\
        & SnapKV                & 30.08 & 41.24 & 56.84 & 56.92 & 47.75 & 29.67 & 24.58 & 24.47 & 24.23 & 64.00 & 92.35 & 41.38 & 7.17 & 99.50 & 64.72 & 57.12 & 47.63 \\    
        & PyramidKV             & 29.50 & 40.46 & 56.47 & 57.30 & 47.55 & 30.34 & 24.26 & 24.46 & 24.00 & 66.50 & 91.32 & 41.64 & 7.20 & 99.50 & 63.65 & 55.49 & 47.48 \\  
        & LAQ                   & 31.64 & 45.55 & 55.21 & 57.73 & 49.60 & 30.99 & 26.67 & 24.79 & 24.85 & 71.00 & 92.33 & 43.06 & 6.92 & 100.00 & 62.16 & 58.45 & 48.81 \\  
        & SpecKV                & 31.12 & 43.77 & 57.22 & 57.51 & 49.32 & 31.06 & 26.34 & 24.61 & 24.90 & 65.00 & 92.13 & 43.32 & 7.00 & 100.00 & 65.31 & 61.89 & 48.78 \\  
        & \textbf{\lookaheadkv}  & 31.39 & 44.92 & 57.56 & 58.56 & 47.72 & 30.82 & 29.24 & 24.82 & 25.83 & 72.50 & 91.92 & 43.39 & 7.08 & 100.00 & 64.87 & 58.36 & \textbf{49.31} \\

        & \multicolumn{18}{c}{\cellcolor{gray!25}\textit{KV Cache Size = 1024}} \\
        & StreamingLLM          & 27.23 & 30.80 & 36.64 & 50.59 & 43.26 & 23.45 & 25.73 & 21.67 & 25.49 & 63.50 & 88.84 & 42.56 & 7.50 & 93.50 & 63.15 & 55.73 & 43.73 \\  
        & SnapKV                & 29.64 & 44.60 & 57.30 & 57.62 & 48.31 & 31.18 & 27.57 & 24.17 & 25.84 & 69.50 & 92.04 & 42.78 & 7.08 & 99.50 & 64.57 & 58.46 & 48.76 \\  
        & PyramidKV             & 30.79 & 44.91 & 56.65 & 58.13 & 48.17 & 30.56 & 26.65 & 24.53 & 25.88 & 68.00 & 91.78 & 42.20 & 6.83 & 99.50 & 64.41 & 57.77 & 48.55 \\  
        & LAQ                   & 31.63 & 45.63 & 55.02 & 57.70 & 50.27 & 31.28 & 28.82 & 25.10 & 26.18 & 72.50 & 92.33 & 43.31 & 6.50 & 100.00 & 62.75 & 59.04 & 49.25 \\  
        & SpecKV                & 31.59 & 45.44 & 57.98 & 57.51 & 49.16 & 31.95 & 28.67 & 24.95 & 25.77 & 67.50 & 92.23 & 43.94 & 6.00 & 99.50 & 65.21 & 62.30 & 49.36 \\  
        & \textbf{\lookaheadkv}  & 31.14 & 46.04 & 57.77 & 58.22 & 48.43 & 30.72 & 30.75 & 25.31 & 26.66 & 72.50 & 91.92 & 43.39 & 7.08 & 100.00 & 64.87 & 58.36 & \textbf{49.57} \\   

        & \multicolumn{18}{c}{\cellcolor{gray!25}\textit{KV Cache Size = 2048}} \\  
        & StreamingLLM          & 28.53 & 37.02 & 39.90 & 51.22 & 45.83 & 23.69 & 28.41 & 21.91 & 26.50 & 67.50 & 90.98 & 42.53 & 7.25 & 90.50 & 64.88 & 57.52 & 45.26 \\  
        & SnapKV                & 31.22 & 46.14 & 56.94 & 58.12 & 48.21 & 31.74 & 30.24 & 24.81 & 26.78 & 71.50 & 91.49 & 43.16 & 6.38 & 99.50 & 64.98 & 58.80 & 49.38 \\  
        & PyramidKV             & 31.37 & 46.01 & 56.61 & 58.02 & 48.21 & 31.50 & 29.73 & 24.70 & 26.57 & 71.50 & 91.65 & 42.83 & 6.64 & 99.50 & 64.94 & 58.32 & 49.26 \\  
        & LAQ                   & 31.30 & 45.69 & 55.62 & 57.61 & 49.91 & 31.33 & 30.96 & 25.51 & 26.77 & 72.50 & 92.33 & 43.54 & 6.83 & 100.00 & 63.77 & 59.28 & 49.56 \\  
        & SpecKV                & 31.88 & 46.64 & 57.39 & 57.97 & 48.80 & 32.72 & 30.96 & 25.38 & 26.82 & 71.00 & 91.48 & 43.65 & 5.88 & 99.50 & 65.79 & 61.16 & \textbf{49.81} \\  
        & \textbf{\lookaheadkv}  & 31.01 & 46.37 & 57.24 & 58.15 & 48.31 & 31.12 & 32.56 & 25.22 & 27.07 & 72.50 & 91.48 & 43.56 & 6.38 & 99.50 & 64.96 & 59.13 & 49.66 \\  
        \cmidrule(lr){2-19}
        \bottomrule
    \end{tabular}
    } 
\end{table*}

\begin{table*}[!ht]
    \centering
    \caption{LongBench evaluation results for Qwen3-8B}
    \label{tab:longbench_qwen8b}
    \resizebox{1.05\linewidth}{!}{%
    \fontsize{20}{20}\selectfont
    \begin{tabular}{llccccccccccccccccc}
        \toprule
        \multicolumn{2}{c}{\multirow{2}{*}{}} & \multicolumn{3}{c}{Single-Document QA} & \multicolumn{3}{c}{Multi-Document QA} & \multicolumn{3}{c}{Summarization} & \multicolumn{3}{c}{Few-shot Learning} & \multicolumn{2}{c}{Synthetic} & \multicolumn{2}{c}{Code} & \multirow{2}{*}{Avg.} \\
        \cmidrule(lr){3-5} \cmidrule(lr){6-8} \cmidrule(lr){9-11} \cmidrule(lr){12-14} \cmidrule(lr){15-16} \cmidrule(lr){17-18}

        & & NrtQA & Qasper & MF-en & HotpotQA & 2WikiMQA & Musique & GovReport & QMSum & MultiNews & TREC & TriviaQA & SAMSum & PCount & Pre & Lcc & RB-P & \\
        \midrule
        \midrule
  
        & FullKV & 26.04 & 47.76 & 53.33 & 59.23 & 43.37 & 36.05 & 33.66 & 24.05 & 24.79 & 71.50 & 90.21 & 44.43 & 2.00 & 100.00 & 69.39 & 65.57 & 49.46 \\

        & \multicolumn{18}{c}{\cellcolor{gray!25}\textit{KV Cache Size = 64}} \\
        & StreamingLLM          & 16.62 & 25.37 & 25.56 & 40.57 & 33.95 & 18.48 & 13.97 & 18.98 & 12.40 & 39.50 & 80.65 & 35.10 & 1.50 & 69.00 & 59.25 & 53.13 & 34.00 \\  
        & SnapKV                & 15.87 & 28.01 & 35.93 & 41.97 & 33.93 & 21.23 & 14.24 & 19.01 & 12.20 & 42.00 & 80.85 & 32.86 & 3.50 & 69.00 & 58.23 & 51.93 & 35.05 \\  
        & PyramidKV             & 15.45 & 27.42 & 36.67 & 42.79 & 34.00 & 19.69 & 14.67 & 18.95 & 12.89 & 43.00 & 80.62 & 33.89 & 2.00 & 71.50 & 58.34 & 52.26 & 35.26 \\  
        & LAQ                   & 16.22 & 32.11 & 45.02 & 42.35 & 37.53 & 21.07 & 15.71 & 19.56 & 13.47 & 44.00 & 76.64 & 35.15 & 3.50 & 83.50 & 59.06 & 52.04 & 37.31 \\  
        & SpecKV                & 15.30 & 27.73 & 44.79 & 37.60 & 36.56 & 12.94 & 16.55 & 19.59 & 14.67 & 32.00 & 58.10 & 32.81 & 4.50 & 74.50 & 60.22 & 54.75 & 33.91 \\  
        & \textbf{\lookaheadkv}  & 22.11 & 43.13 & 51.85 & 59.01 & 42.50 & 34.34 & 22.66 & 21.61 & 18.49 & 64.50 & 88.75 & 39.48 & 1.50 & 67.75 & 62.78 & 56.07 & \textbf{43.53} \\   
        
        & \multicolumn{18}{c}{\cellcolor{gray!25}\textit{KV Cache Size = 128}} \\
        & StreamingLLM          & 17.65 & 26.69 & 28.40 & 41.05 & 33.46 & 20.82 & 15.72 & 19.15 & 15.14 & 43.00 & 82.57 & 38.44 & 1.50 & 70.00 & 62.86 & 56.69 & 35.82 \\
        & SnapKV                & 19.14 & 32.65 & 45.99 & 54.81 & 38.95 & 26.59 & 17.66 & 20.83 & 16.04 & 49.50 & 87.10 & 38.90 & 3.50 & 99.50 & 64.62 & 58.29 & 42.13 \\
        & PyramidKV             & 15.57 & 30.19 & 41.84 & 46.01 & 35.73 & 19.57 & 16.51 & 19.67 & 14.86 & 47.00 & 83.51 & 35.56 & 2.50 & 92.00 & 62.14 & 53.07 & 38.48 \\
        & LAQ                   & 22.74 & 42.15 & 53.55 & 57.89 & 42.84 & 36.74 & 21.33 & 22.25 & 18.34 & 64.50 & 89.55 & 40.93 & 3.00 & 100.00 & 66.74 & 61.70 & 46.52 \\
        & SpecKV                & 23.03 & 37.14 & 53.58 & 56.77 & 42.24 & 31.82 & 21.33 & 22.86 & 19.04 & 60.00 & 88.31 & 41.50 & 3.50 & 100.00 & 66.82 & 61.96 & 45.62 \\
        & \textbf{\lookaheadkv}  & 26.06 & 44.30 & 53.24 & 58.78 & 42.79 & 35.89 & 25.29 & 22.95 & 21.13 & 66.50 & 88.95 & 41.64 & 3.50 & 99.50 & 65.95 & 62.88 & \textbf{47.46} \\ 

        & \multicolumn{18}{c}{\cellcolor{gray!25}\textit{KV Cache Size = 256}} \\
        & StreamingLLM          & 18.18 & 28.53 & 28.52 & 42.81 & 33.58 & 21.34 & 18.63 & 19.20 & 17.76 & 48.00 & 85.58 & 40.08 & 1.00 & 69.00 & 65.50 & 59.41 & 37.32 \\
        & SnapKV                & 23.03 & 38.32 & 51.04 & 57.36 & 40.67 & 32.82 & 21.51 & 21.89 & 18.97 & 59.50 & 89.46 & 41.06 & 2.00 & 100.00 & 67.62 & 61.88 & 45.45 \\
        & PyramidKV             & 18.47 & 34.87 & 47.44 & 55.68 & 37.89 & 26.67 & 20.43 & 20.92 & 17.43 & 58.50 & 85.20 & 38.98 & 3.50 & 100.00 & 65.51 & 57.32 & 43.05 \\
        & LAQ                   & 26.00 & 45.44 & 53.84 & 57.00 & 43.53 & 36.62 & 24.22 & 23.38 & 20.38 & 70.00 & 89.05 & 42.47 & 3.00 & 100.00 & 68.17 & 64.03 & 47.95 \\
        & SpecKV                & 22.58 & 41.09 & 53.89 & 59.85 & 42.42 & 34.50 & 24.53 & 23.64 & 21.25 & 68.00 & 88.13 & 43.12 & 3.00 & 100.00 & 68.39 & 64.40 & 47.42 \\
        & \textbf{\lookaheadkv}  & 25.88 & 45.40 & 52.68 & 58.47 & 44.05 & 36.13 & 27.77 & 23.71 & 22.88 & 69.00 & 89.05 & 43.32 & 2.00 & 100.00 & 67.83 & 64.71 & \textbf{48.31} \\  

        & \multicolumn{18}{c}{\cellcolor{gray!25}\textit{KV Cache Size = 512}} \\
        & StreamingLLM          & 18.94 & 30.86 & 30.21 & 43.89 & 33.26 & 22.51 & 22.24 & 19.62 & 21.16 & 58.50 & 87.48 & 41.11 & 2.00 & 57.00 & 67.06 & 61.59 & 38.59 \\  
        & SnapKV                & 24.63 & 43.72 & 51.96 & 58.37 & 42.36 & 34.04 & 25.03 & 22.55 & 21.66 & 69.00 & 89.53 & 42.06 & 3.00 & 100.00 & 69.37 & 64.96 & 47.64 \\  
        & PyramidKV             & 23.12 & 40.52 & 51.43 & 57.57 & 40.89 & 32.85 & 23.85 & 21.92 & 19.70 & 68.50 & 89.55 & 40.89 & 3.00 & 100.00 & 67.47 & 61.73 & 46.44 \\  
        & LAQ                   & 27.34 & 46.98 & 53.70 & 57.31 & 43.35 & 37.64 & 26.93 & 23.67 & 22.19 & 72.50 & 88.96 & 43.81 & 3.00 & 100.00 & 68.23 & 64.71 & 48.77 \\  
        & SpecKV                & 24.22 & 45.65 & 54.34 & 60.53 & 43.85 & 35.26 & 27.21 & 24.04 & 22.53 & 70.50 & 90.20 & 43.71 & 3.50 & 100.00 & 69.25 & 65.85 & 48.79 \\  
        & \textbf{\lookaheadkv}  & 25.33 & 46.49 & 52.04 & 59.32 & 43.09 & 36.92 & 29.56 & 23.80 & 24.01 & 71.50 & 90.21 & 44.20 & 2.00 & 100.00 & 68.88 & 65.58 & \textbf{48.93} \\  

        & \multicolumn{18}{c}{\cellcolor{gray!25}\textit{KV Cache Size = 1024}} \\
        & StreamingLLM          & 21.25 & 32.82 & 31.44 & 45.94 & 34.38 & 23.34 & 25.73 & 20.25 & 23.50 & 62.00 & 88.71 & 41.18 & 0.50 & 44.00 & 68.39 & 63.65 & 39.19 \\
        & SnapKV                & 24.26 & 46.13 & 52.48 & 58.52 & 42.66 & 36.89 & 28.39 & 23.61 & 23.33 & 69.00 & 89.55 & 43.13 & 2.00 & 100.00 & 69.05 & 66.27 & 48.45 \\
        & PyramidKV             & 23.77 & 42.89 & 53.01 & 58.86 & 42.32 & 35.47 & 27.32 & 23.07 & 22.72 & 71.00 & 89.95 & 42.56 & 2.00 & 100.0 & 68.81 & 64.25 & 48.00 \\
        & LAQ                   & 26.11 & 47.27 & 53.45 & 57.01 & 43.52 & 37.26 & 29.50 & 23.88 & 23.47 & 71.50 & 89.63 & 44.00 & 2.00 & 100.00 & 67.94 & 64.83 & 48.84 \\
        & SpecKV                & 24.98 & 46.56 & 54.07 & 59.04 & 43.37 & 34.12 & 29.32 & 24.18 & 23.68 & 71.00 & 90.11 & 44.56 & 3.00 & 100.00 & 69.09 & 66.53 & 48.98 \\
        & \textbf{\lookaheadkv}  & 25.36 & 47.23 & 52.56 & 59.30 & 43.25 & 36.39 & 31.65 & 23.72 & 24.61 & 71.00 & 90.21 & 44.69 & 0.50 & 100.00 & 68.93 & 65.22 & \textbf{49.04} \\

        & \multicolumn{18}{c}{\cellcolor{gray!25}\textit{KV Cache Size = 2048}} \\  
        & StreamingLLM          & 21.73 & 38.54 & 38.02 & 47.96 & 36.78 & 25.53 & 28.69 & 21.47 & 24.11 & 65.00 & 90.30 & 42.85 & 1.00 & 48.50 & 68.37 & 64.63 & 41.47 \\  
        & SnapKV                & 25.55 & 47.52 & 53.20 & 58.73 & 42.70 & 36.08 & 30.64 & 23.78 & 24.40 & 71.50 & 90.21 & 43.27 & 1.10 & 100.00 & 69.33 & 65.36 & 48.96 \\  
        & PyramidKV             & 25.47 & 46.69 & 53.21 & 58.41 & 42.90 & 36.61 & 29.41 & 23.61 & 24.18 & 71.50 & 90.05 & 42.90 & 1.10 & 100.00 & 69.28 & 65.21 & 48.78 \\  
        & LAQ                   & 24.94 & 47.22 & 53.71 & 57.72 & 43.45 & 37.40 & 31.24 & 24.03 & 24.51 & 72.50 & 90.13 & 44.54 & 2.00 & 100.00 & 68.29 & 64.68 & 49.15 \\  
        & SpecKV                & 24.86 & 47.00 & 53.80 & 61.43 & 43.74 & 34.94 & 31.38 & 23.98 & 24.57 & 70.50 & 91.11 & 44.46 & 0.00 & 100.00 & 69.12 & 65.87 & 49.17 \\  
        & \textbf{\lookaheadkv}  & 26.76 & 48.01 & 52.92 & 59.43 & 43.20 & 36.21 & 32.64 & 23.93 & 24.93 & 71.00 & 90.21 & 44.74 & 1.00 & 100.00 & 69.23 & 65.01 & \textbf{49.33} \\   
        \cmidrule(lr){2-19}
        \bottomrule
    \end{tabular}
    } 
\end{table*}

\newpage

\subsection{Results on RULER}
\begin{figure}[h]
    \centering
    \adjincludegraphics[width=0.99\linewidth, max width=0.99\columnwidth]{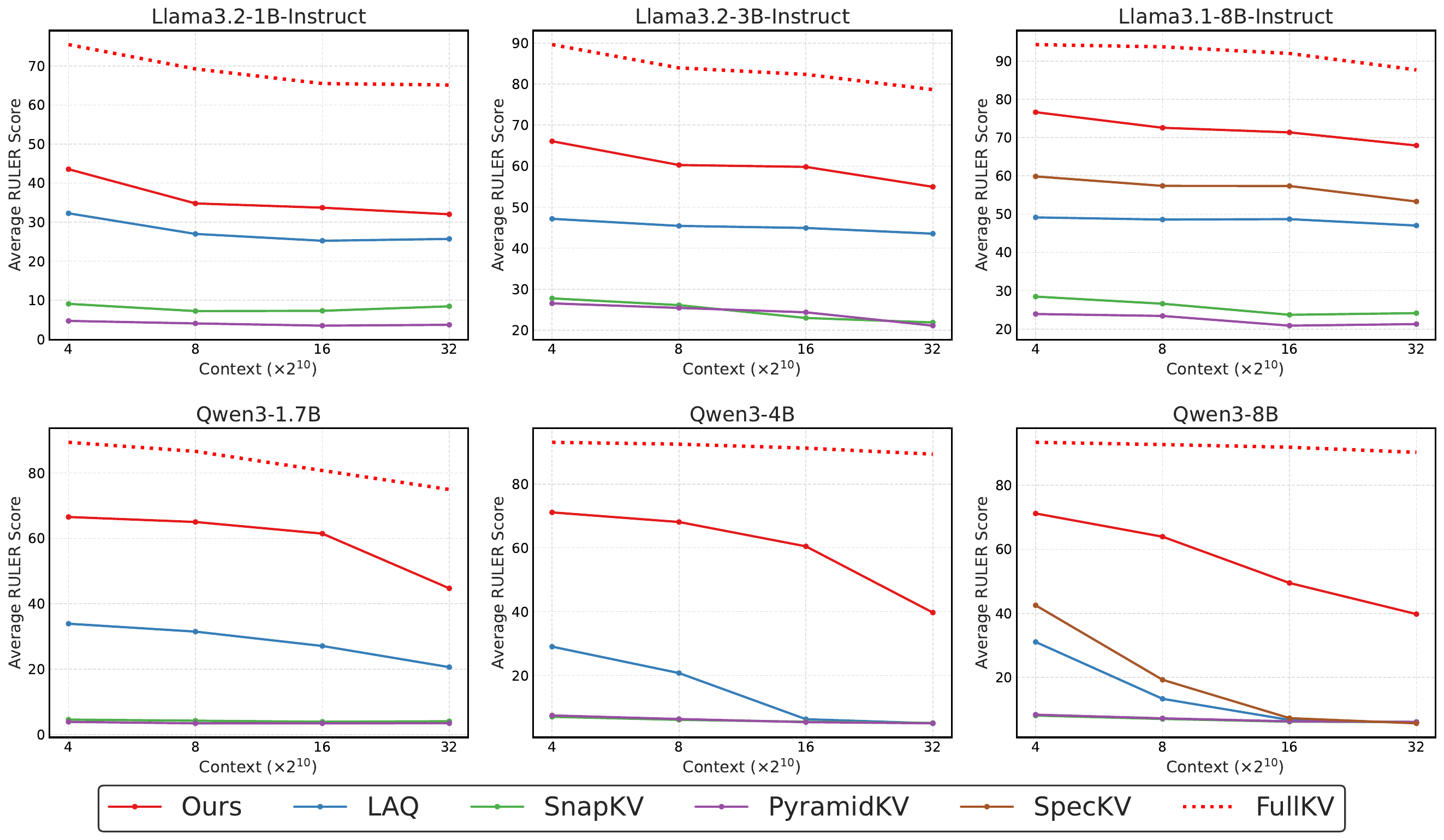}  
    \caption{
    Full RULER results across context lengths (budget = $64$)
    }\vspace{-0.3cm}
    \label{fig:ruler_main_64}
\end{figure}

\begin{figure}[h]
    \centering
    \adjincludegraphics[width=0.99\linewidth, max width=0.99\columnwidth]{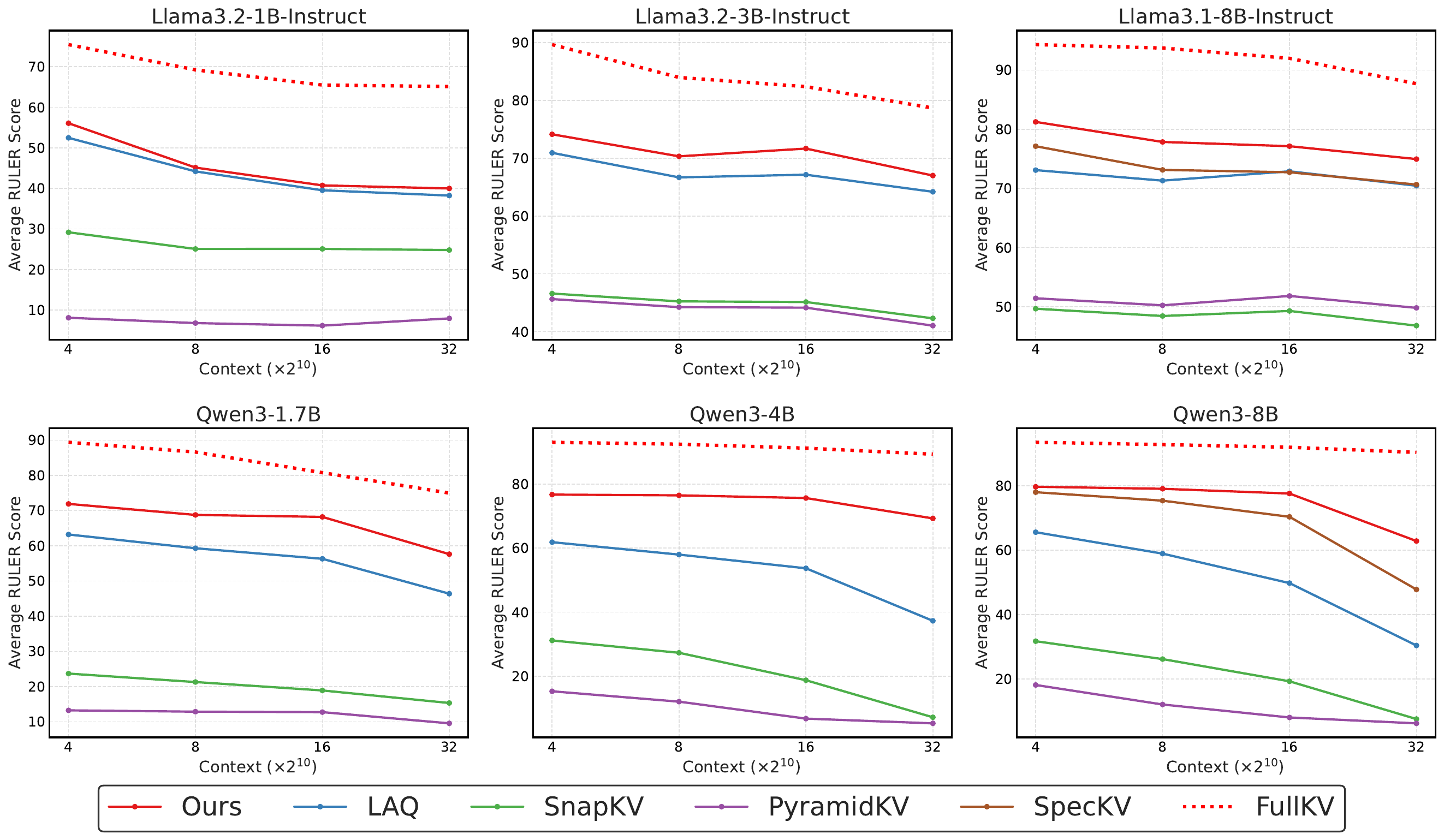}  
    \caption{
    Full RULER results across context lengths (budget = $128$)
    }\vspace{-0.3cm}
    \label{fig:ruler_main_128}
\end{figure}

We report the RULER results across all six models tested, with cache budget settings at 64 (\cref{fig:ruler_main_64}) and 128 (\cref{fig:ruler_main_128}).
\newpage

\subsection{Additional Efficiency Analysis}
We show the full results of the latency analysis that were omitted in the main paper due to space limitations in this section. Note that the empirical TTFT overheads for some methods can be larger than theoretical estimations. These are probably due to a combination of measurement noise and inefficient implementation of these methods in KVCache-Factory or their official implementations. Better implementations may reduce these overheads significantly, more in line with the theoretical cost.
\begin{table}[H]
  \centering
  \caption{Theoretical and Practical Analysis across various context lengths and methods.}
  \label{tab:kv-theoretical-practical-full}
  \begingroup
    \setlength{\tabcolsep}{4pt}       
    \renewcommand{\arraystretch}{1.35} 

    \resizebox{\textwidth}{!}{%
      \begin{tabular}{c l S[table-format=4, round-mode=places, round-precision=0] S[table-format=3, round-mode=places, round-precision=0] S[table-format=4, round-mode=places, round-precision=0] S[table-format=3.2, round-mode=places, round-precision=2] S[table-format=4, round-mode=places, round-precision=0] S[table-format=3.2, round-mode=places, round-precision=2]}
        \toprule
        & & \multicolumn{4}{c}{\textbf{Theoretical Cost}} & \multicolumn{2}{c}{\textbf{Empirical Cost}} \\
        \cmidrule(lr){3-6}\cmidrule(lr){7-8}
        \shortstack[c]{\textbf{Context}\\\textbf{Length}} &
        \textbf{Method} &
        \shortstack[c]{\textbf{Compute}\\\textbf{(TFLOPs)}} &
        \shortstack[c]{\textbf{Memory Traffic}\\\textbf{(GB)}} &
        \shortstack[c]{\textbf{TTFT}\\\textbf{(ms)}} &
        \shortstack[c]{\textbf{TTFT}\\\textbf{Overhead (ms)}} &
        \shortstack[c]{\textbf{TTFT}\\\textbf{(ms)}} &
        \shortstack[c]{\textbf{TTFT}\\\textbf{Overhead (ms)}} \\
        \midrule
        \multirow{5}{*}{$4$K}
          & Forward Pass Only     & 60.00  & 13.00  & 113.37 & N/A       & 130.08    & N/A      \\
          & \lookaheadkv    & 60.47  & 13.05  & 114.30 & 0.9164    & 141.46 & 11.3800    \\
          & SnapKV                & 60.00  & 13.02  & 113.38 & 0.0085    & 143.22 & 13.1400    \\
          & SpecKV                & 69.79  & 76.94  & 165.48 & 52.1031 & 222.50 & 92.4200   \\
          & LAQ                   & 60.94  & 443.52 & 347.19 & 233.8089 & 636.66 & 506.5800  \\
        \midrule
        \multirow{5}{*}{$8$K}
          & Forward Pass Only     & 136.00 & 13.00  & 256.99 & N/A      & 290.81  & N/A          \\
          & \lookaheadkv    & 136.53 & 13.05  & 258.03   & 1.0345 & 301.69  & 10.8800   \\
          & SnapKV                & 136.00 & 13.02  & 257.00   & 0.0085 & 310.98  & 20.1700  \\
          & SpecKV                & 159.12 & 81.07  & 336.52 & 79.5264 & 411.32 & 120.5100  \\
          & LAQ                   & 137.06 & 444.52 & 491.58 & 234.5877 & 800.19 & 509.3800 \\
        \midrule
        \multirow{5}{*}{$16$K}
          & Forward Pass Only     & 336.00 & 13.00  & 634.92 & N/A       & 658.32  & N/A        \\
          & \lookaheadkv    & 336.66 & 13.05  & 636.19   & 1.2707 & 676.82 & 18.5000   \\
          & SnapKV                & 336.00 & 13.02  & 634.93   & 0.0085 & 695.44  & 37.1200   \\
          & SpecKV                & 397.78 & 89.32  & 791.97 & 157.0489 & 865.63 & 207.3100  \\
          & LAQ                   & 337.31 & 446.52 & 871.07 & 236.1451 & 1181.86 & 523.5400  \\
        \midrule
        \multirow{5}{*}{$32$K}
          & Forward Pass Only     & 928.00 & 13.00  & 1753.59& N/A      & 1760.22  & N/A      \\
          & \lookaheadkv    & 928.91 & 13.05  & 1755.33   & 1.7431 & 1798.26   & 38.0400   \\
          & SnapKV                & 928.00 & 13.02  & 1753.60   & 0.0085 & 1837.89  & 77.6700   \\
          & SpecKV                & 1115.09& 105.82 & 2156.39 & 402.7969 & 2263.09  & 502.8700  \\
          & LAQ                   & 929.81 & 450.52 & 1992.85 & 239.2601 & 2313.90 & 553.6800  \\
        \bottomrule
      \end{tabular}%
    }
  \endgroup
\end{table}
\newpage

\section{Hyper-Parameters}
\textbf{Training Hyper-parameters.}
\label{app:hyperparams}
Learning rate was searched for Llama and Qwen model family among [\num{5e-5}, \num{1e-4}, \num{2e-4}, \num{1e-3}]. The final hyper-parameters for all experiments are shown in \cref{hparams_llama}.

\begin{table}[H]
\begin{center}
\caption{Training hyperparameters.}
\begin{tabular}{lc}
\hline
\toprule
\textbf{Parameters} & \textbf{Values} \\
\toprule
Optimizer & Adam \\
$\beta_1, \beta_2$ & $0.9, 0.95$ \\
Effective Batch Size & $32$ \\
Drop-out ($p$) & $0.0$ \\
Max Sequence Length & $16384$ (prompt length) + $512$ (response length) \\
Train Iters & $7600$ \\
Learning rate & \num{1e-3} (for Llama), \num{2e-4} (for Qwen) \\
Schedule & Cosine \\

Warmup steps & $2\%$  \\ 
Min LR & $0.0$ \\
Gradient clipping & $1.0$ \\
\bottomrule
\end{tabular}
\label{hparams_llama}
\end{center}
\end{table}

\textbf{Eviction Hyper-parameters.}
We use the implementations in KVCache-Factory or their official implementations (SpecKV) for all baseline methods, except for LAQ which we re-implement ourselves due to the lack of an official release. Following prior works~\citep{li2024snapkv, cai2024pyramidkv, galim2025speckv}, we use standard configuration settings for all baseline methods, including an observation window size of $32$, maxpooling kernel size of 7, and mean reduction for GQA compatibility~\citep{feng2024ada}. For \lookaheadkv, we use the same settings, except we do not use window size, as our method does not train with the suffix window for prediction. Further, since our lookahead size \(n_{\text{lookahead}}\) is $32$, we set the maximum generation limit of LAQ and SpecKV to $32$ tokens so that the methods can be compared using the same number of draft tokens.

\section{Datasets, Benchmarks, and Software}
\label{app:urls}

\textbf{Software} Our source code is available in the supplementary, and our implementation is built on \href{https://github.com/Zefan-Cai/KVCache-Factory}{KVCache-Factory}.

\textbf{Training Dataset}
Our training dataset mixture consist of random samples from publicly available datasets: 50K long\_sft subset of \href{https://huggingface.co/datasets/nvidia/ChatQA2-Long-SFT-data}{ChatQA2-Long-SFT-data}, 20K subset of \href{https://huggingface.co/datasets/allenai/tulu-3-sft-olmo-2-mixture}{tulu-3-sft-olmo-2-mixture}, 7K samples from \href{https://huggingface.co/datasets/bigcode/the-stack}{The Stack}, and 3K samples from \href{https://huggingface.co/datasets/abacusai/MetaMathFewshot}{MetaMathFewshot}, \href{https://huggingface.co/datasets/abacusai/HellaSwag_DPO_FewShot}{HellaSwag\_DPO\_Fewshot}, and \href{https://huggingface.co/datasets/abacusai/ARC_DPO_FewShot}{ARC\_DPO\_Fewshot}, respectively.

\textbf{Evaluation Benchmarks} We used LongBench dataset as fetched and processed by KVCache-Factory, see \href{https://huggingface.co/datasets/zai-org/LongBench}{HF Dataset} for the official source. For RULER, we used \href{https://github.com/NVIDIA/RULER}{RULER Github}. For LongProc, we used \href{https://github.com/princeton-pli/LongProc}{LongProc Github}.

\section{LLM Usage}
LLM assistants were used to refine the wording of selected sentences, while the majority of the text was written by human. All LLM‑generated text was carefully inspected to ensure that it contained no harmful or controversial content. Additionally, we used LLMs to help in finding some of the related literature discussed in the paper.

\end{document}